%% file: bare_jrnl_new_sample4.tex
\documentclass[lettersize,journal]{IEEEtran}
\usepackage{amsmath,amsfonts}
\usepackage{algorithmic}
\usepackage{algorithm}
\input{preamble}

\usepackage[pagebackref,breaklinks,colorlinks,allcolors=blue]{hyperref}
\usepackage{booktabs}
\usepackage{multirow}
\usepackage{array}
\usepackage[caption=false,font=normalsize,labelfont=sf,textfont=sf]{subfig}
\usepackage{textcomp}
\usepackage{stfloats}
\usepackage{url}
\usepackage{verbatim}
\usepackage{graphicx}
\usepackage{cite}
\usepackage{booktabs}
\usepackage{multirow}
\usepackage{makecell}
\usepackage[table]{xcolor}
\newcommand{\rev}[1]{#1}
\newcommand{\revMinor}[1]{#1}
\begin{document}

\title{UniReg: Conditional Unified Model for Medical Image Registration}

\author{Zi~Li,~\IEEEmembership{Student Member,~IEEE}, Jianpeng~Zhang, Tai~Ma, Tony~C.~W.~Mok, Yan-Jie~Zhou, Zeli~Chen, Xianghua~Ye, Le~Lu,~\IEEEmembership{Fellow,~IEEE}, Cheng~Chen,~\IEEEmembership{Member,~IEEE}, and Dakai~Jin,~\IEEEmembership{Member,~IEEE}
\thanks{ Z. Li and C. Chen are with the Department of Electrical and Computer Engineering, The University of Hong Kong, Hong Kong SAR, China.
J. Zhang, T. Ma, T. Mok, Y. Zhou, Z. Chen, and D. Jin are with DAMO Academy, Alibaba Group.
X. Ye is with The First Affiliated Hospital, Zhejiang University School of Medicine, Hangzhou, China.
L. Lu is with Ant Group.}
\thanks{ Corresponding authors: J. Zhang (e-mail: jianpeng.zhang0@gmail.com) and C. Chen (e-mail:  cchen@eee.hku.hk).}
}

\markboth{Journal of \LaTeX\ Class Files,~Vol.~14, No.~8, August~2021}%
{Shell \MakeLowercase{\textit{et al.}}: A Sample Article Using IEEEtran.cls for IEEE Journals}


\maketitle

\input{sec/0_abstract}


\input{sec/1_intro}

\input{sec/2_relatedwork}

\input{sec/3_method}

\input{sec/4_experiment}

\input{sec/5_conclusion}

\section*{Acknowledgments}
This work was supported by the Zhejiang Provincial ``Spearhead \& Pathfinder + X'' R\&D Breakthrough Program under Grant No. 2024C03043, the National Natural Science Foundation of China under Grant No. 82573722, and the Zhejiang Provincial Natural Science Foundation of China under Grant No. 2024-KYI-00I-I05.

\bibliographystyle{IEEEtran}
\bibliography{IEEEabrv,refs}

\input{sec/biography}

\end{document}

%% file: preamble.tex
\usepackage[accsupp]{axessibility}  
\usepackage{graphicx}
\usepackage{amsmath}
\usepackage{amssymb}
\usepackage{booktabs}
\usepackage{times}
\usepackage{epsfig}

\DeclareSymbolFont{matha}{OML}{txmi}{m}{it}
\DeclareMathSymbol{\varv}{\mathord}{matha}{118}
\usepackage{array}
\usepackage{tabularx}
\usepackage{multirow, booktabs}
\usepackage{float}
\usepackage{bm}
\usepackage{resizegather}

\usepackage{pifont}
\usepackage{fontawesome}
\usepackage{tikz}

\usepackage{caption}

\def\F{\mathbf{F}}
\def\M{\mathbf{M}}

\def\E{\mathbf{E}}

%% file: sec/0_abstract.tex
\begin{abstract}
Learning-based medical image registration has matched the accuracy of conventional methods while offering superior computational efficiency. However, existing approaches suffer from poor generalization across diverse clinical scenarios, requiring the laborious development of multiple isolated networks for specific registration tasks, \emph{e.g.}, inter-/intra-subject registration or anatomical region-specific alignment, leading to cumbersome development pipelines. 
To overcome this limitation, we propose \textbf{UniReg}, the first conditional unified model for multi-scenario \revMinor{medical} image registration, which combines the precision advantages of task-specific learning methods with the generalization of traditional optimization methods. 
Our key innovation is a unified registration framework that adaptively estimates deformation fields conditioned on: (1) anatomical structure priors, (2) registration type constraints (inter/intra-subject), and (3) instance-specific features, 
enabling effective alignment across heterogeneous \revMinor{CT and MR registration} scenarios within a single model.
Through comprehensive experiments \rev{on multiple CT/MR registration datasets}, UniReg achieves \rev{superior average registration accuracy} compared with current state-of-the-art learning-based methods while exhibiting strong cross-scenario generalization. 
Moreover, by replacing multiple isolated task-specific models with a compact unified model, UniReg substantially reduces the overall training burden in terms of total training cost and model redundancy.
\end{abstract}

%% file: sec/1_intro.tex
\section{Introduction}
\label{sec:intro}

\begin{figure}[!t]
	\begin{center}
        \includegraphics[width=1.0\linewidth]{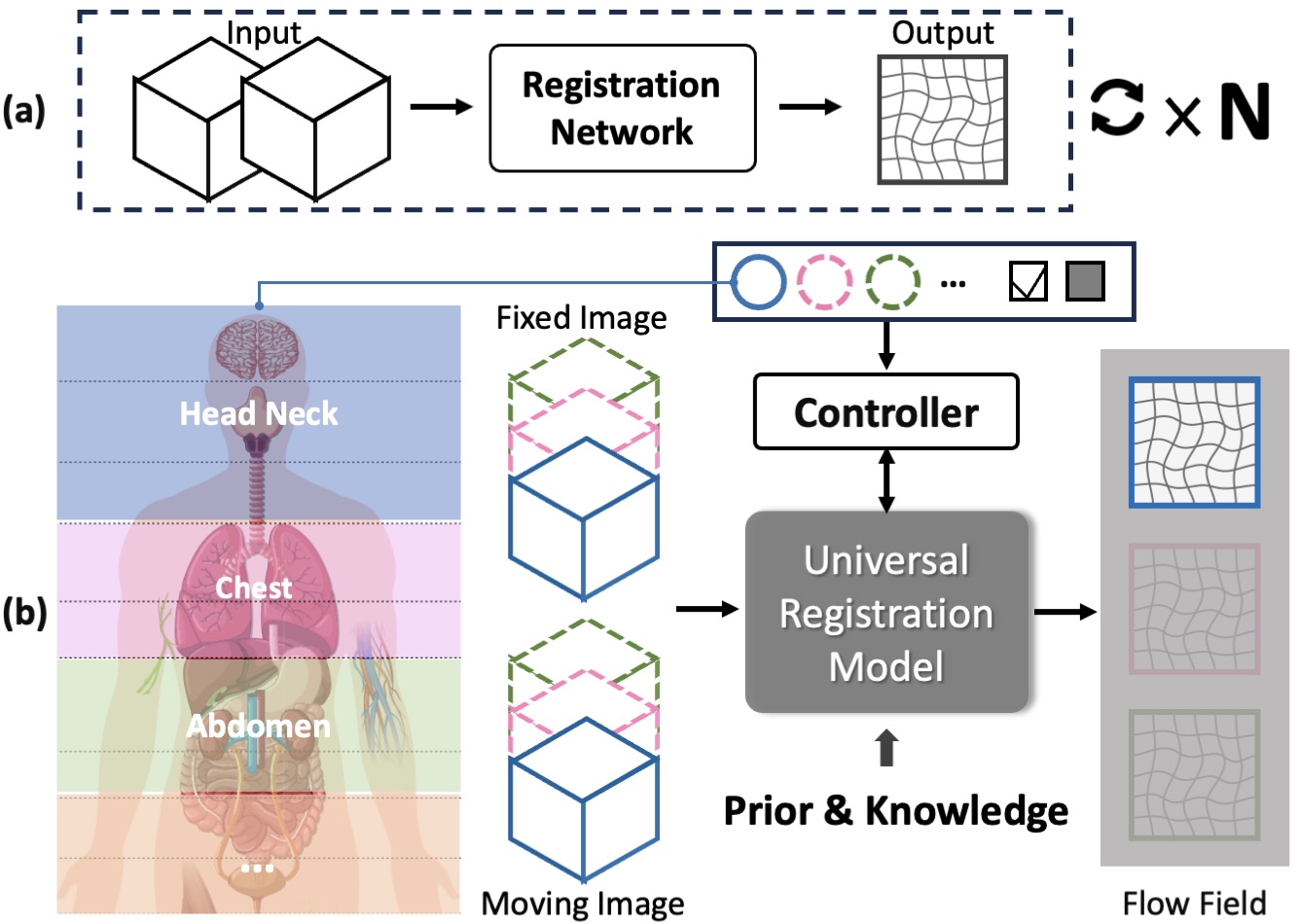}
	\end{center}
 \vspace{-0.2em} 
    \caption{\textbf{(a)} Task-specific registration model: Conventional learning-based registration requires manual design of specialized networks for each anatomical region and registration type. This results in redundant development efforts, where $N$ distinct tasks necessitate training $N$ isolated models. \textbf{(b)} One universal registration model: UniReg introduces a conditional control module that dynamically adapts to anatomical regions (circle symbols), registration types (rectangle symbols: inter/intra-subject), and imaging characteristics. Through systematic integration of anatomical priors and knowledge within a unified architecture, the model achieves whole-body organ registration with task-aware deformation field generation.}
	\label{fig:motivation}
\end{figure}

\IEEEPARstart{E}{stablishing} accurate anatomical correspondence~\cite{maurer1993review} across \revMinor{medical} images is paramount for medical image analysis and its associated clinical applications, encompassing atlas-based segmentation~\cite{ZhaoBDGD19,LiLLLF22,li2025leveraging}, diagnostic procedures~\cite{alam2018medical,hu2025ai}, and longitudinal studies~\cite{QinBSPPNR18,ezhil2009determination}.

Conventional image registration approaches~\cite{SotirasDP13,SunNK14,HeinrichJBS12,AvantsTSCKG11,shen2002hammer,pluim2000image} are often used to facilitate automatic image registration for a wide variety of registration tasks. Yet, conventional approaches are slow in practice as they rely on an iterative numerical optimization framework. These approaches estimate the optimal deformation by gradually maximizing the similarity between the target image and the transformed reference image in an iterative fashion. Iterative optimization in conventional methods leads to a heavy computation burden, prohibiting their usage in high-throughput environments that need to process dozens to thousands of image scans. 

Learning-based registration methods~\cite{BalakrishnanZSG19,MokC20,LiuLFZHL22,hu2022cross,bigalke2023unsupervised,mok2024modality} addressed the inefficiency of conventional methods by reformulating the iterative optimization problem to a learning problem using convolutional neural networks. Numerous studies explored advanced network architectures~\cite{MokC20,meng2022non,ma2024iirp,meng2024correlation} and implemented more efficient feature extractors~\cite{LiTMBWGZLYYJ23,LiuYHGLYHXXYJ21,mok2024modality} to improve precision.

However, due to the learning nature of these methods, the training and registration processes are highly customized and task-specific, leading to inferior generalizability and flexibility compared to conventional approaches. A trivial approach to adapting a learning-based approach to multi-scenario registration tasks is to develop multiple isolated registration networks and train them on independent task-specific datasets, as shown in Figure~\ref{fig:motivation}\textbf{(a)}, which requires massive computational resources and tedious hyperparameter tuning to maximize the registration performance for each task. Alternatively, one can develop a \revMinor{unified} registration model by training a registration network over various datasets. Yet, this approach has proved ineffective and unstable \cite{TianGKVEBRN24} as there are significant variations in the optimal set of registration hyperparameters across different image modalities, anatomical regions, and tasks. Training a typical convolutional neural network to learn multiple diverse registration tasks often results in sub-optimal registration accuracy.

In recent years, pioneering works of deformable registration \cite{mok2021conditional,hoopes2021hypermorph,DeyRDG21,AbulnagaHDHFGD25} introduce conditional image registration frameworks that are able to model the effect of a wide range of regularization hyperparameters on the deformation field. While these conditional image registration frameworks offer enhanced flexibility in hyperparameter tuning, the registration network remains constrained to a single task or anatomical region.

\subsection{Contributions}
In this paper, we present a \revMinor{conditional \textbf{Uni}fied registration framework for multi-scenario medical image} \textbf{Reg}istration, referred to as UniReg, as depicted in Figure~\ref{fig:motivation}\textbf{(b)}, representing a significant advancement towards bridging the gap between registration performance and generalization. The model comprises a shared backbone and a conditioning control mechanism. The shared backbone is dedicated to learning robust and generalized anatomical correspondences, while the conditioning control module dynamically generates optimal deformation fields for diverse scenarios, demonstrating a high level of adaptability. 
To initiate this conditioning control process, we encode various scenarios by integrating task-specific information, such as anatomical structures, inter- and intra-subject variations, and body parts. Moreover, registration images are employed as part of the conditioning vector, enabling the model to provide optimal solutions tailored to each registration image. We then develop a lightweight controller network that synthesizes these variable factors and encodes them into a task-specific deformation kernel, which is subsequently transmitted to the dynamic registration head to generate the deformation field.

Furthermore, our UniReg model is trained on a wide variety of \rev{CT/MR} registration scenarios and datasets, enhancing its applicability across diverse contexts and showcasing strong generalization capabilities. Moreover, \rev{we introduce task-specific optimization priors, including similarity-window and regularization settings, to accommodate different deformation characteristics across anatomical regions and registration types.}
Specifically, our datasets encompass inter-subject registration tasks across multiple anatomical regions, including the head-neck, chest, and abdomen CT scans. They also include intra-subject registration tasks derived from multiphase liver CT scans, lung respiratory dynamics datasets, \rev{and cardiac MR images, as well as inter-subject brain MR registration}. The CT datasets \rev{alone} feature a diverse array of $90$ anatomical structures, representing nearly the entirety of the human body, from the cranial region to the pelvic area, while also incorporating tumor-affected patients (liver/lung tumors). 
The detailed dataset statistics are summarized in Table~\ref{tab:dataset}.

Empirically, UniReg exhibits \rev{competitive and often superior performance compared with contemporary state-of-the-art registration methods across diverse CT/MR registration scenarios (Section~\ref{sec:mainresults})}. By replacing multiple independently trained task-specific models with a single conditional unified model, UniReg substantially reduces the required training iterations, which translates to a substantial decrease in the overall training cost.
The main contributions are as follows:
\begin{itemize}
 \item 
 We propose the first conditional unified registration framework (UniReg) that features a single network that dynamically adapts to various tasks. It eliminates the need for developing multiple task-specific registration networks, substantially reduces computational costs, and lessens reliance on expertise in model development.

 \item
 We introduce a conditional control vector for \revMinor{unified} registration that encodes anatomical structure priors, registration type constraints, and instance-specific features. A dynamic deformation generation module is further developed that facilitates adaptive registration using dynamic kernels. These components collectively provide an efficient and streamlined solution for registration tasks across diverse domains.
 
 \item  
 We design a dynamic training strategy that leverages prior knowledge and human expertise to optimize the UniReg model. This strategy incorporates regularization priors and anatomical insights to tailor the optimization process for each specific registration task, enhancing the registration performance.

 \item 
 UniReg demonstrates strong flexibility and efficacy in handling complex anatomical variations and diverse registration scenarios. The comprehensive evaluation covers \rev{multiple CT and MR registration tasks, involving a wide range of anatomical structures and tumors from the cranial region to the pelvic area, together with a landmark-based fine-grained correspondence assessment.} To the best of our knowledge, this constitutes the most extensive assessment of its kind to date.
\end{itemize}

%% file: sec/2_relatedwork.tex
\section{Related Work}

\subsection{Deformable Image Registration}
Traditional deformable registration methods~\cite{SunNK14,HeinrichJBS12,heinrich2013towards,AvantsTSCKG11,shen2002hammer,pluim2000image,avants2009advanced} formulate image registration as an optimization problem and iteratively minimize an image similarity measure, often together with a deformation regularizer, to align a pair of images. 
Recently, learning-based deformable registration methods~\cite{BalakrishnanZSG19,MokC20,mok2020fast,ChenFHSLD22,bigalke2023unsupervised,mok2022unsupervised,LiuLZFL20,mok2024modality,wang2024improving} have employed deep networks to directly predict a displacement vector field or to estimate a velocity field from which the transformation is obtained via integration. Compared with optimization-based approaches, learning-based methods are substantially faster during inference.
\rev{Some studies have further improved registration efficiency through compact or efficient architectures~\cite{Chen2026EOIR,hu2022cross}, while others have developed dense registration modules for misaligned multi-modal image fusion~\cite{Zheng2025PlugAndPlay,wang2024improving}.}

To better capture large deformations, recent methods have also explored more sophisticated CNN or Transformer architectures. 
\revMinor{These strategies~\cite{ZhaoDCX19,LiuLFZHL22,MokC20,hu2022cross,FanLLWLLH23,mok2023deformable,ma2023pivit,meng2022non,hu2022recursive,zhou2023self} }commonly rely on multi-step estimation, pyramidal representations, cascaded structures, iterative refinement, or the integration of registration with image synthesis tasks~\cite{xu2019deepatlas}.
\revMinor{For example, the studies in~\cite {hu2022recursive,zhou2023self} improve deformable registration through recursive decomposition and self-distilled hierarchical refinement, respectively.}
In addition, registration performance can be improved by incorporating anatomical information during training through segmentation labels~\cite{BalakrishnanZSG19,xu2019deepatlas,zhang2025voxelopt}. More recently, SAME and SAMConvex~\cite{LiTMBWGZLYYJ23,LiuYHGLYHXXYJ21} have investigated the use of pre-trained Self-supervised Anatomical eMbeddings~\cite{yan2022sam} as registration features, which provide discriminative anatomical semantic information.

Despite these advances, most existing methods are still designed and optimized for specific registration tasks or application scenarios. When transferred to a new anatomical region, imaging modality, or registration type, the optimization procedure or network weights typically need to be re-executed or retrained. 
\revMinor{By contrast, our work focuses on a unified framework that can be readily applied across different anatomical regions, imaging modalities, and registration scenarios.}

\subsection{Conditional Medical Image Registration}
Conditional image registration has drawn considerable research attention in the community due to its potential in amortized hyperparameter learning \cite{mok2021conditional,hoopes2021hypermorph}. Mok~\emph{et al.}~\cite{mok2021conditional} designed a learning paradigm for conditional image registration, in which the registration network is conditioned on the dynamic regularization strength with linear modulation \cite{huang2017arbitrary}. 
Dey~\emph{et al.}~\cite{DeyRDG21} introduced a generative adversarial registration framework conditioned on flexible image covariates for template estimation. 
Hoopes~\emph{et al.}~\cite{hoopes2021hypermorph} proposed to learn the effects of registration hyperparameters on the deformation field with Hypernetworks \cite{ha2016hypernetworks}, which leverage a secondary network to generate the conditioned weights for the entire network layers. This design is more flexible than traditional convolutional networks, where hyperparameters are fixed during training and inference. 

Beyond the registration domain, several approaches have focused on enhancing the representational capacity and flexibility of segmentation networks through dynamic filters. 
Jia~\emph{et al.}~\cite{jia2016dynamic} designed a dynamic filter network in which the filters are generated dynamically and conditioned on the input image. Subsequently, Zhang~\emph{et al.}~\cite{zhang2021dodnet} introduced the conditionally parameterized convolutions, which learn task-specific convolution kernels for each assigned segmentation task.

\subsection{Unified Medical Image Registration}
Despite a vast number of methods for deformable image registration, most of these methods are task-specific \cite{LiuLFZHL22,MokC20,FanLLWLLH23} or require manual~\cite{SunNK14,HeinrichJBS12} or automated~\cite{hoopes2021hypermorph,mok2021conditional,LiuLFZHL22} hyperparameter tuning for new registration tasks. 

To address this issue, some attempts have been made to explore unified medical image registration. 
Siebert \emph{et al.} \cite{siebert2024convexadam} proposed a self-configuring dual-optimization method, which amortizes the hyperparameter search using a combination of rule-based and grid search techniques. 
Hoffmann \emph{et al.} \cite{hoffmann2021synthmorph} proposed a deep learning-based registration method that learns from randomly synthesized images and shapes. While learning from synthesized shapes circumvents the task-specific limitation, it lacks anatomical knowledge of the human structures, resulting in sub-optimal registration accuracy and plausibility. 
Tian~\emph{et al.}\cite{TianGKVEBRN24} collected multiple datasets from different medical domains and co-trained a unified medical image registration network capable of multi-task registration.
However, it employs a uniform regularization approach that fails to incorporate essential task-specific priors. 

In contrast to these works, our conditional unified model realizes dynamic registration through a dedicated controller mechanism that integrates rich task-specific knowledge and anatomical priors, thereby enabling flexible and adaptive solutions for multi-scenario image registration.

%% file: sec/3_method.tex
\section{UniReg Method}\label{sec:formatting}

\begin{figure*}[!ht]
	\begin{center}
        \includegraphics[width=0.85\linewidth]{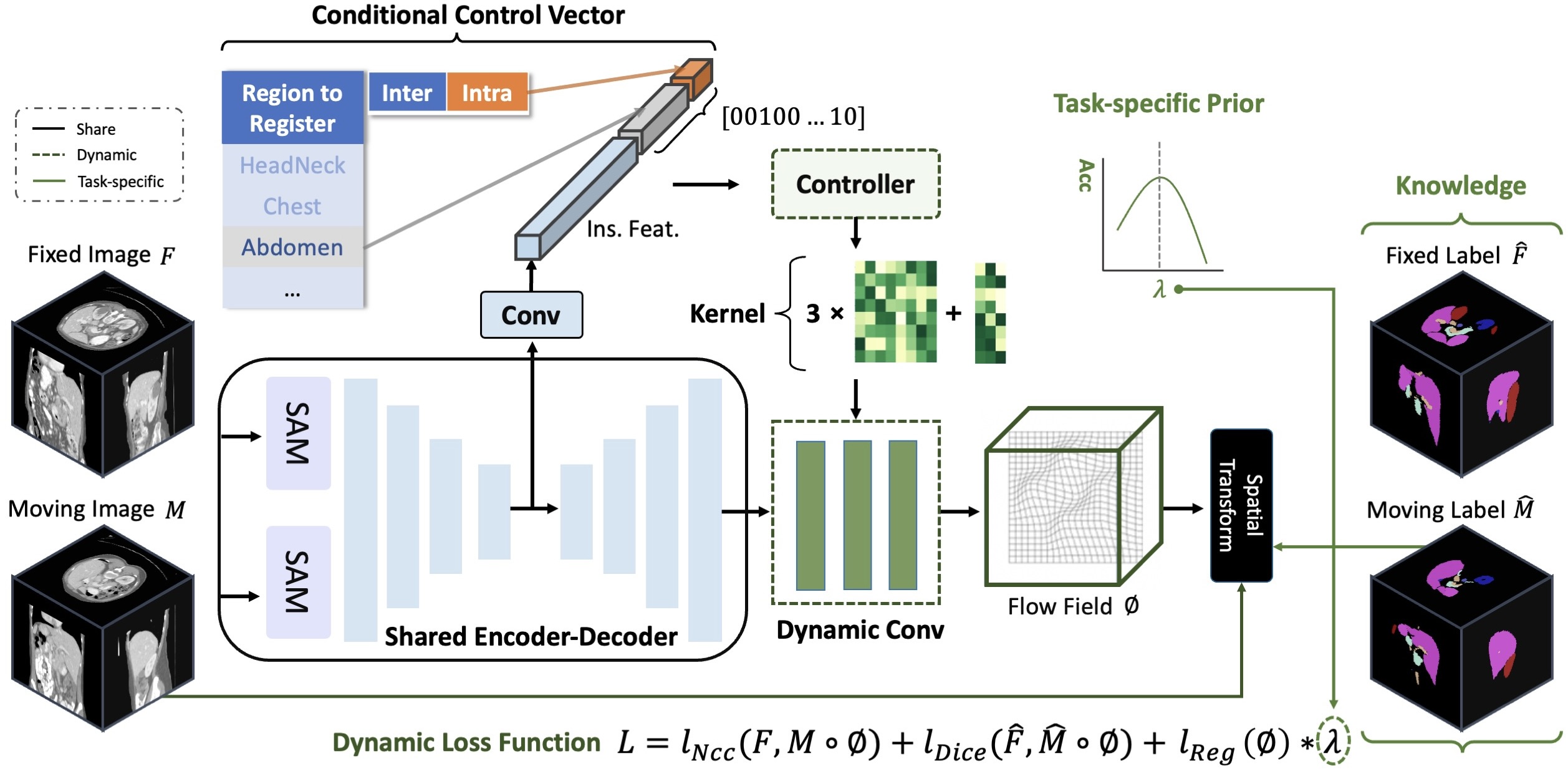}
	\end{center}
    \caption{\textbf{Overview of the UniReg framework.} Our framework comprises a backbone architecture and dynamic registration. Leveraging our dynamic controller mechanism, we can incorporate human expertise and prior knowledge of diverse registration tasks throughout the training process. At each training iteration, our framework selects task-specific hyperparameters \rev{corresponding to distinct task IDs, thereby instituting optimization objectives tailored to the individual registration task. When anatomical annotations are available, optional segmentation-based constraints can also be activated for semi-supervised training.}}
	\label{fig:overview}
\end{figure*}

\subsection{Problem Formulation}
Consider $n$ registration datasets $\{\mathcal{D}_1, \mathcal{D}_2, ..., \mathcal{D}_n\}$ from $n$ imaging scenarios. 
Here, $\mathcal{D}_i = \{\F_{ij}, \M_{ij}, \hat{\F}_{ij}, \hat{\M}_{ij}\}^{m_i}_{j=1}$ represents the $i$-th dataset that contains $m_i$ image pairs. Specifically, $\F_{ij}$, $\M_{ij}$ are fixed and moving volumes defined over a three-dimensional spatial domain $\Omega \subseteq \mathbb{R}^{D \times W \times H}$, where $W \times H$ is the spatial size of each slice and $D$ is the number of slices. 
\rev{When available,} the corresponding anatomical labels are denoted as $\hat{F}_{ij}$ and $\hat{M}_{ij}$.
The core of registration is a mapping function that learns a transformation to align the moving image with the fixed image space. 
To address the $n$ different registration tasks, conventional solutions necessitate optimizing $n$ registration models, with individual parameters ${\theta_1, \theta_2, ..., \theta_n}$, to handle each task individually, formulated as follows:
\begin{equation}\label{eq:funtions}
\begin{cases}
  \ \min_{\theta_1} \sum_{j=1}^{m_1} \ \mathcal{L}_1(f_1(\F_{1j}, \M_{1j};  \theta_1),  \hat{\F}_{1j}, \hat{\M}_{1j}) \\
  \ \qquad \vdots \\
  \  \min_{\theta_n} \sum_{j=1}^{m_n} \ \mathcal{L_n}(f_n(\F_{nj}, \M_{nj};  \theta_n),  \hat{\F}_{nj}, \hat{\M}_{nj}) 
\end{cases}
\end{equation}
where $\mathcal{L}$ refers to the loss function of registration networks. \rev{Here, $\hat{\F}$ and $\hat{\M}$ are optional annotations.}
In contrast, we attempt to process all registration tasks using one single network $f$, \emph{i.e.}, UniReg, formulated as
\begin{equation}\label{eq:ourfuntion}
{\min}_{\theta,\psi} {\sum_{i=1}^{n}} {\sum_{j=1}^{m_i}}  \mathcal{L}(f(\F_{ij}, \M_{ij}; \theta, h(i; \psi))).
\end{equation}

\subsection{Conditional Unified Model}
We first introduce the basic structure of UniReg in Section~\ref{sec:SharedBackbone} and then describe how we define the conditioning vector in Section~\ref{sec:Vector}, which enables the dynamic application of UniReg to optimally estimate parameters for each registration scenario in Section~\ref{sec:DynamicDeformation}. We elaborate on our training and detail several extra considerations during inference in Section~\ref{sec:train}.

\subsubsection{Shared Backbone}\label{sec:SharedBackbone}
The backbone is designed to generate features that encompass the semantic information of multiple organs and is not constrained to a specific task. 
For this purpose, our unified encoder-decoder architecture is \rev{based on the Self-supervised Anatomical embedding (SAM)~\cite{yan2022sam}}, serving as the basic framework of our registration model, as illustrated in the left part of Figure~\ref{fig:overview}. This architecture facilitates both global and local feature extraction from image pairs, shown to effectively enhance registration~\cite{LiTMBWGZLYYJ23,LiuYHGLYHXXYJ21}.
Subsequently, our encoder architecture is structured as a series of repeated residual blocks, each comprising two convolutional layers. Each convolutional layer is succeeded by an activation function. 
In each downsampling phase, a convolutional operation characterized by a stride of $2$ is employed, thereby reducing the resolution of the input feature maps by half.
Symmetrically, the decoder progressively upsamples the feature map to enhance its resolution. At every stage, the upsampled feature map is combined with the corresponding low-level feature map from the encoder. Now, we have $\E_{ij} = f(\F_{ij}, \M_{ij}; \theta)$, where $\theta$ represents all backbone parameters.
Here, $\E_{ij} \in \mathbb{R}^{C\times D \times W \times H}$ and $C$ is the channel dimension. 

\rev{Notably, UniReg is a flexible and backbone-agnostic registration framework, rather than being restricted to a specific encoder-decoder architecture. To demonstrate its generality, we further evaluate two UniReg variants using coarse-to-fine CNN~\cite{ma2024iirp} and coarse-to-fine MLP~\cite{meng2024correlation} backbones.}

\subsubsection{Conditional Control Vector}\label{sec:Vector}

\rev{The task-aware control mechanism conditions deformation generation on anatomical task identity, registration type, and instance-specific image context. 
For the $l$-th dynamic deformation head, let 
$\mathbf{H}_{ij}^{l}\in\mathbb{R}^{C_l\times D_l\times W_l\times H_l}$ 
denote the backbone-derived feature map used for dynamic flow prediction. 
The default shared-backbone instantiation contains a single deformation head ($l=1$), which predicts the flow at one resolution, whereas the coarse-to-fine CNN and MLP instantiations use $l\in\{1,2,3,4\}$ to denote deformation heads at different pyramid levels.
We first aggregate the spatial information of $\mathbf{H}_{ij}^{l}$ by global average pooling (GAP):
\begin{equation}
    \mathbf{g}_{ij}^{l}=\operatorname{GAP}(\mathbf{H}_{ij}^{l})
    \in\mathbb{R}^{C_l}.
\end{equation}}

\rev{The anatomical task identity is encoded as an $n$-dimensional one-hot vector:
\begin{equation}
    \mathbf{t}_{i,k}=
    \begin{cases}
    1, & k=i,\\
    0, & k\neq i,
    \end{cases}
    \quad k=1,2,\dots,n,
\end{equation}
where $\mathbf{t}_{i,k}=1$ indicates that the current registration pair belongs to the $i$-th anatomical task.}

\rev{We encode the registration type as a two-dimensional one-hot vector:
\begin{equation}
    \mathbf{r}_{ij,k} =
    \begin{cases}
    1, & k=\tau_{ij},\\
    0, & k\neq \tau_{ij},
    \end{cases}
    \quad k=1,2,
\end{equation}
where $\tau_{ij}=1$ denotes inter-subject registration and $\tau_{ij}=2$ denotes intra-subject registration.
}

\rev{The final conditional vector for the $l$-th controller is defined as:
\begin{equation}
    \mathbf{z}_{ij}^{l}
    =
    [\mathbf{g}_{ij}^{l}\Vert \mathbf{t}_{i}\Vert \mathbf{r}_{ij}]
    \in\mathbb{R}^{C_l+n+2},
\end{equation}
where $\Vert$ denotes vector-level concatenation. 
This formulation avoids direct concatenation between dense feature maps and low-dimensional task vectors.}

\subsubsection{Dynamic Deformation Generation}\label{sec:DynamicDeformation}

\rev{Inspired by adaptive architectures~\cite{jia2016dynamic, zhang2021dodnet}, we use a lightweight controller to generate task-aware parameters for the deformation head, as shown in the right part of Figure~\ref{fig:overview}. 
Given the conditional vector $\mathbf{z}_{ij}^{l}$, the $l$-th controller is formulated as
\begin{equation}
    \bm{\omega}_{ij}^{l}
    =
    \mathcal{G}^{l}(\mathbf{z}_{ij}^{l};\theta_{\mathcal{G}}^{l}),
\end{equation}
where $\theta_{\mathcal{G}}^{l}$ denotes the learnable parameters of the controller, and $\bm{\omega}_{ij}^{l}$ contains the weights and biases of three cascaded $1\times1\times1$ convolutional layers:
\begin{equation}
    \bm{\omega}_{ij}^{l}
    =
    \{\mathbf{W}^{l,1}_{ij},\mathbf{b}^{l,1}_{ij},
    \mathbf{W}^{l,2}_{ij},\mathbf{b}^{l,2}_{ij},
    \mathbf{W}^{l,3}_{ij},\mathbf{b}^{l,3}_{ij}\}.
\end{equation}}

\rev{The dynamic deformation head predicts a displacement field as
\begin{equation}
\begin{aligned}
    \mathbf{Q}_{ij}^{l,1}
    &=
    \sigma\left(\mathbf{W}^{l,1}_{ij} * \mathbf{H}_{ij}^{l}
    + \mathbf{b}^{l,1}_{ij}\right),\\
    \mathbf{Q}_{ij}^{l,2}
    &=
    \sigma\left(\mathbf{W}^{l,2}_{ij} * \mathbf{Q}_{ij}^{l,1}
    + \mathbf{b}^{l,2}_{ij}\right),\\
    \Delta\phi_{ij}^{l}
    &=
    \mathbf{W}^{l,3}_{ij} * \mathbf{Q}_{ij}^{l,2}
    + \mathbf{b}^{l,3}_{ij},
\end{aligned}
\end{equation}
where $*$ denotes $1\times1\times1$ convolution and $\sigma(\cdot)$ denotes LeakyReLU. 
The output $\Delta\phi_{ij}^{l}\in\mathbb{R}^{3\times D_l\times W_l\times H_l}$ represents the displacement field predicted at the $l$-th stage.}



\begin{table*}[hb]
\centering
\setlength{\tabcolsep}{7pt}
\renewcommand{\arraystretch}{1.12}

\textbf{(a) CT datasets}

\vspace{0.3em}

\resizebox{0.85\textwidth}{!}{%
\begin{tabular}{lcccccccc}
\toprule
\multirow{2}{*}{\textbf{Dataset}} 
& \multirow{2}{*}{\textbf{Task}} 
& \multirow{2}{*}{\textbf{Modality}} 
& \multirow{2}{*}{\textbf{\# Label}} 
& \multicolumn{3}{c}{\textbf{\revMinor{No. Image pairs}}} 
& \multirow{2}{*}{\textbf{\revMinor{Original size}}} 
& \multirow{2}{*}{\textbf{BPR range}} \\
\cmidrule(lr){5-7}
& & & & \textbf{Train} & \textbf{Val} & \textbf{Test} & & \\
\midrule
HeadNeck & Inter-subject & Mono  & 40 & 4446 & 90 & 812 & $280\times280\times192$ & 0.51--1.00 \\
Chest    & Inter-subject & Mono  & 35 & 4160 & 90 & 90  & $180\times180\times200$ & 0.31--0.84 \\
Abdomen  & Inter-subject & Mono  & 13 & 380  & 0  & 90  & $250\times250\times160$ & 0.00--0.35 \\
Lung GTV & Intra-subject & Mono  & 1  & 0    & 0  & 35  & $256\times256\times112$ & 0.60--0.65 \\
Liver    & Intra-subject & Multi & 2  & 150  & 10 & 50  & $290\times290\times160$ & 0.33--0.53 \\
\rev{DIR-QA} & Intra-subject & Mono & 60 & 0 & 0 & 60 & $224\times224\times192$ & 0.00--0.35 \\
\bottomrule
\end{tabular}%
}

\vspace{0.9em}

\textbf{\rev{(b) MR datasets}}

\vspace{0.3em}

\resizebox{0.75\textwidth}{!}{%
\begin{tabular}{lccccccc}
\toprule
\multirow{2}{*}{\textbf{Dataset}} 
& \multirow{2}{*}{\textbf{Task}} 
& \multirow{2}{*}{\textbf{Modality}} 
& \multirow{2}{*}{\textbf{\# Label}} 
& \multicolumn{3}{c}{\textbf{\revMinor{No. Image pairs}}} 
& \multirow{2}{*}{\textbf{\revMinor{Original size}}} \\
\cmidrule(lr){5-7}
& & & & \textbf{Train} & \textbf{Val} & \textbf{Test} & \\
\midrule
\rev{Cardiac} & Intra-subject & Mono & 3 & 180 & 20 & 100 &  $128\times128\times16$\\
\rev{LUMIR}   & Inter-subject & Mono & 0 & 11.44M & 0 & 0 & $160\times224\times192$ \\
\rev{OASIS}   & Inter-subject & Mono & 35 &  0 & 19 & 500 &  $160\times224\times192$\\
\bottomrule
\end{tabular}%
}

\caption{\rev{Statistics of the CT and MR datasets used in our experiments.}}
\label{tab:dataset}
\end{table*}

\begin{table*}[hb]
\centering
\setlength{\tabcolsep}{3.5pt}
\renewcommand{\arraystretch}{1.15}
\fontsize{8.4pt}{10.1pt}\selectfont
\begin{tabular}{lcccccccc}
\toprule
\textbf{Task} 
& \textbf{Task ID} 
& \textbf{Sampling weight} 
& \textbf{Learning rate} 
& \textbf{Sim. loss}
& \textbf{NCC win.}
& \textbf{Sim. weight}
& \textbf{Reg. loss}
& \textbf{Reg. weight} \\
\midrule
Brain (Stage 1)
& 5 
& 1 
& $1.0\times10^{-4}$ 
& NCC
& $9\times9\times9$
& 1.0
& Grad-$L_2$
& 1.0 \\

Brain (Stage 2) 
& 5 
& 2 
& $1.0\times10^{-5}$ 
& NCC
& $9\times9\times9$
& 1.0
& Grad-$L_2$
& 1.0 \\

Chest 
& 0 
& 2 
& $1.0\times10^{-5}$ 
& NCC
& $9\times9\times9$
& 1.0
& Grad-$L_2$
& 1.0 \\

Abdomen 
& 1 
& 2 
& $1.0\times10^{-5}$ 
& NCC
& $9\times9\times9$
& 1.0
& Grad-$L_2$
& 1.0 \\

HeadNeck 
& 2 
& 2 
& $1.0\times10^{-5}$ 
& NCC
& $9\times9\times9$
& 1.0
& Grad-$L_2$
& 1.0 \\

Liver 
& 3 
& 2 
& $1.0\times10^{-5}$ 
& NCC
& $9\times9\times9$
& 1.0
& Grad-$L_2$
& 10.0 \\

ACDC 
& 4 
& 4 
& $1.0\times10^{-5}$ 
& NCC
& $3\times9\times9$
& 1.0
& Grad-$L_2$
& 0.5 \\

\bottomrule
\end{tabular}
\caption{\rev{Training settings for the six-task joint training of UniReg. NCC denotes the normalized cross-correlation image similarity loss, and Grad-$L_2$ denotes the gradient-based deformation regularization loss.}}
\label{tab:six_task_training_settings}
\end{table*}

\subsection{Training and Inference}\label{sec:train}

Conceptually, the loss function of UniReg is defined as:
\begin{equation}
\begin{array}{l}
 \! \mathcal{L} \! = \! \mathcal{L}_{Sim}(\F, \M \!\circ \! \phi ) +  \lambda \mathcal{L}_{Reg}(\phi) \! 
\end{array}
\end{equation}
where a similarity term $\mathcal{L}_{Sim}$ penalizes the appearance differences. The regularizer $\mathcal{L}_{Reg}$ promotes spatial smoothness within the transformation map. 
The operator $\circ$ denotes the warping operation implemented by a spatial transformer. 
Here, $\lambda$ is a task-specific trade-off hyperparameter.

\subsubsection{Regularization Priors}
Considering that the precision of the resultant transformation map is significantly affected by the selection of specific hyperparameter values~\cite{mok2021conditional,hoopes2021hypermorph,LiuLFZHL22}, we pre-establish task-related hyperparameter priors $\lambda = \{\lambda_1, \lambda_2, \dots, \lambda_n\}$ using \cite{mok2021conditional}. Therefore, during the training process, the strength of regularization will be calibrated according to the task condition, ensuring the model understands the exact objectives it needs to achieve and further allowing it to handle multiple tasks.


\subsubsection{Optional Anatomical Knowledge}
\rev{UniReg can optionally incorporate voxel-wise anatomical segmentation masks as additional spatial priors. This optional design provides a semi-supervised instantiation of UniReg, in which region-specific segmentation constraints can be selectively activated to further guide deformation estimation. }

\subsubsection{Inference}
During inference, UniReg can be flexibly applied to diverse registration tasks. Given a test image pair, anatomical region information, and registration type, the backbone network extracts image features, while the controller generates task- and instance-conditioned kernels based on the corresponding control vector. The dynamic head then uses these kernels to predict the deformation field and register the task-specified anatomical structures.

%% file: sec/4_experiment.tex
\begin{table*}[t]
\centering
\setlength{\tabcolsep}{4.2pt}
\renewcommand{\arraystretch}{1.18}
\definecolor{hl}{RGB}{245,222,222}

\resizebox{\textwidth}{!}{%
\begin{tabular}{llcccccccccccc}
\toprule[1.5pt]
\multicolumn{2}{c}{\textbf{Method}} 
& \multicolumn{2}{c}{\textbf{HeadNeck CT}} 
& \multicolumn{2}{c}{\textbf{Chest CT}} 
& \multicolumn{2}{c}{\textbf{Abdomen CT}} 
& \multicolumn{2}{c}{\textbf{Liver CT}} 
& \multicolumn{2}{c}{\textbf{Cardiac}} 
& \multicolumn{2}{c}{\textbf{Brain}} \\
\cmidrule(lr){3-4}
\cmidrule(lr){5-6}
\cmidrule(lr){7-8}
\cmidrule(lr){9-10}
\cmidrule(lr){11-12}
\cmidrule(lr){13-14}
& 
& \textbf{DSC$_{40}$}$\uparrow$ & \textbf{SDlogJ}$\downarrow$
& \textbf{DSC$_{35}$} & \textbf{SDlogJ}
& \textbf{DSC$_{13}$} & \textbf{SDlogJ}
& \textbf{DSC$_2$} & \textbf{$\%(|J_{\phi}|<0)$}$\downarrow$
& \textbf{DSC$_3$} & \revMinor{\textbf{$\%(|J_{\phi}|<0)$}}
& \textbf{DSC$_{35}$} & \textbf{SDlogJ} \\
\midrule[1.pt]

\multirow{10}{*}{\textbf{\shortstack{Task-\\specific}}}


& Initial
& 46.72$^{\dagger}$ & 1.04
& 45.14$^{\dagger}$ & 0.08
& 40.14$^{\dagger}$ & 1.87
& 78.05$^{\dagger}$ & 0.00
& 58.14$^{\dagger}$ & 0.00
& 56.57$^{\dagger}$ & 0.00 \\


& NiftyReg
& 52.92$^{\dagger}$ & 0.16
& 51.58$^{\dagger}$ & 0.04
& 34.98$^{\dagger}$ & 0.22
& 81.92$^{\dagger}$ & 0.01
& 63.97$^{\dagger}$ & 0.00
& 78.23$^{\dagger}$ & 1.11 \\

& Deeds
& 54.21$^{\dagger}$ & 7.18
& 52.72$^{\dagger}$ & 1.28
& 46.52$^{\dagger}$ & 0.44
& 83.50$^{\dagger}$ & 5.81
& 75.01$^{\dagger}$ & 0.02
& 73.84$^{\dagger}$ & 3.84 \\

& \rev{RAN}
& 47.03$^{\dagger}$ & 0.94
& 46.10$^{\dagger}$ & 1.98
& 46.17$^{\dagger}$ & 1.30
& 82.48$^{\dagger}$ & 0.01
& 71.48$^{\dagger}$ & 0.00
& 79.76$^{\dagger}$ & 0.01 \\

& \rev{TransMorph}
& 47.26$^{\dagger}$ & 1.17
& 53.52$^{\dagger}$ & 7.11
& 46.24$^{\dagger}$ & 0.85
& 82.26$^{\dagger}$ & 0.01
& 74.97$^{\dagger}$ & 0.61
& 80.66$^{\dagger}$ & 0.01 \\

& LapIRN
& 48.61$^{\dagger}$ & 0.03
& 55.87 & 4.33
& 46.44$^{\dagger}$ & 0.72
& 81.02$^{\dagger}$ & 0.01
& 71.43$^{\dagger}$ & 0.00
& 75.43$^{\dagger}$ & 0.15 \\

& IIRP-Net
& 51.97$^{\dagger}$ & 1.72
& 54.41$^{\dagger}$ & 3.03
& 52.19$^{\dagger}$ & 5.46
& 82.08$^{\dagger}$ & 0.18
& 75.37 & 0.06
& 81.77 & 0.01 \\

& CorrMLP
& 50.35$^{\dagger}$ & 4.87
& 54.40$^{\dagger}$ & 4.42
& 53.03$^{\dagger}$ & 7.28
& 79.91$^{\dagger}$ & 0.29
& 77.31 & 0.10
& 81.89 & 0.01 \\

& SAME
& 48.60$^{\dagger}$ & 1.08
& 55.36 & 0.40
& 49.27$^{\dagger}$ & 2.82
& 80.64$^{\dagger}$ & 0.03
& \multicolumn{2}{c}{N/A}
& \multicolumn{2}{c}{N/A} \\
\midrule[1.pt]

\multirow{3}{*}{\textbf{Unified}}
& uniGradICON
& \multicolumn{2}{c}{---}
& 47.92$^{\dagger}$ & 0.54
& 48.30$^{\dagger}$ & 0.31
& \multicolumn{2}{c}{---}
& \textbf{79.56}$^{\dagger}$ & 0.01
& 78.77$^{\dagger}$ & 1.28 \\

& UniReg (CNN)
& 48.33 & 0.91
& 54.45 & 2.31
& 47.12 & 2.51
& 83.96 & 0.03
& \multicolumn{2}{c}{N/A}
& \multicolumn{2}{c}{N/A} \\

& \rev{UniReg (C2F)}
& \textbf{57.04} & 0.01
& \textbf{56.69} & 0.01
& 55.08 & 0.01
& \textbf{87.08} & 0.00
& 76.10 & 0.08
& 81.79 & 0.01 \\

& \rev{UniReg (MLP)}
& 57.02 & 0.01
& 56.48 & 0.01
& \textbf{55.32} & 0.01
& 87.03 & 0.00
& 76.21 & 0.12
& \textbf{81.97} & 0.01 \\

\bottomrule[1.5pt]
\end{tabular}%
}
\caption{Quantitative comparison between task-specific and unified registration models. The subscript of each DSC metric indicates the number of anatomical structures involved. $\uparrow$: higher is better, and $\downarrow$: lower is better. 
\revMinor{Statistical significance is assessed on DSC using t-test between each comparison method and UniReg (C2F)}, with significant differences ($P<0.05$) marked by $^{\dagger}$.
The best-performing results are shown in bold.
\revMinor{``---'' indicates failed results, inferior to the initial DSC, while ``N/A'' denotes settings that are not applicable. }
}
\label{tab:main_result}
\end{table*}

\section{Experimental Results and Analysis}

\subsection{Dataset and Pre-processing}\label{sec:dataformulation}

The divisions of training, validation, and testing for each registration task are summarized in Table~\ref{tab:dataset}, which also presents other key statistics for all datasets. 
To deepen the understanding of the distribution of body parts within each dataset, we include the body part regression (BPR)~\cite{YanLS18} axial ranges of the CT images.
All data were anonymized and used under appropriate ethics approval. We included:

\noindent\textit{Inter-subject task on HeadNeck.}
We utilized a CT dataset~\footnote{https://segrap2023.grand-challenge.org/} consisting of 240 head-neck cancer patients. The dataset includes annotations for 40 organs.

\noindent\textit{Inter-subject task on Chest.}
A chest CT dataset comprising 94 subjects was collected from hospitals. Each chest CT image features 35 manually labeled anatomical structures identified by a senior radiologist.

\noindent\textit{Inter-subject task on Abdomen.}
The Abdomen CT dataset from Learn2Reg~\footnote{https://learn2reg.grand-challenge.org/Learn2Reg2020/} includes 30 scans, each containing 13 manually labeled anatomical structures.

\noindent\textit{Intra-subject task on Liver.}
We gathered a multi-phase contrast-enhanced liver CT dataset of 80 tumor-diagnosed patients, with scans taken in three phases: pre-contrast, arterial, and venous. A senior radiologist annotated and verified the liver and tumor masks.

\noindent\textit{Intra-subject task on Lung.}
We collected a 4D CT dataset of lungs comprising 35 patients, each featuring paired inspiratory and expiratory breath-hold images. 
The primary gross tumor volume (GTV) in each image was annotated by a senior radiologist.

\noindent\rev{\textit{Intra-subject landmark-based task on Abdomen.}
We included the Abdominal DIR-QA dataset~\cite{criscuolo2025vessel} to assess fine-grained anatomical correspondence. It contains 30 abdominal CT image pairs with more than 1,800 vessel-bifurcation landmarks. We evaluated both registration directions, resulting in 60 directed registration pairs.}

\noindent\rev{\textit{Inter-subject task on Brain MRI.}
We trained the model on all 3,383 unlabeled LUMIR~\cite{dufumier2022openbhb,taha2023magnetic,marcus2007open} MR scans. Since dense labels are unavailable for the full LUMIR dataset, we used the fully labeled OASIS dataset~\cite{marcus2007open,SiebertHH21} for validation and testing, where each MR scan is annotated with 35 anatomical labels. Specifically, the official 19 OASIS validation pairs were used for validation. For testing, we generated 500 fixed random non-identical directed pairs from the remaining labeled cases.}

\noindent\rev{\textit{Intra-subject task on Cardiac MRI.}
We used the ACDC~\cite{bernard2018deep}, including 90 training, 10 validation, and 50 test subjects. Each subject contains end-diastolic (ED) and end-systolic (ES) images with manual segmentations of the left ventricular blood pool, myocardium, and right ventricle. Bidirectional ED-to-ES and ES-to-ED registration pairs were constructed.}

\textit{Pre-processing.}
\rev{For CT datasets, } all images were resampled to an isotropic spacing of $2\times2\times2$ mm, oriented to the RAI direction, and adjusted to a size of $192\times192\times160$ through padding or cropping. 
For inter-subject tasks, SAMCoarse~\cite{abs-2311-14986} was applied for pre-alignment to standardize the evaluation, whereas affine registration was not included for intra-subject tasks due to the minimal linear misalignment. 
\rev{For MR datasets, we directly resized the brain MR images to $160\times224\times192$ and the ACDC cardiac MR images to $128\times128\times16$.}  All images were normalized to the range of $[-1, +1]$.

\subsection{Evaluations, Baselines and Implementation Details}

\subsubsection{Evaluation Metrics}
To quantify registration performance, we register each image pair, propagate the anatomical segmentation map using the predicted transformation, and measure the segmentation overlap within the region of interest using the Dice similarity coefficient (DSC)~\cite{dice1945measures}. 
\rev{We additionally report the target registration error (TRE), a landmark-based metric.}
To evaluate the plausibility of the predicted deformation fields, we compute the percentage of voxels with a negative Jacobian determinant ($\%(|J_{\phi}|<0)$), where a negative determinant indicates local folding~\cite{Ashburner07}. 
We also report the standard deviation of the logarithm of the Jacobian determinant (SDlogJ) as an additional measure of deformation regularity. 
In our experiments, SDlogJ is used for inter-subject registration, while the folding percentage is used for intra-subject registration.

\begin{figure}[t]
	\centering
    \includegraphics[width=0.85\linewidth]{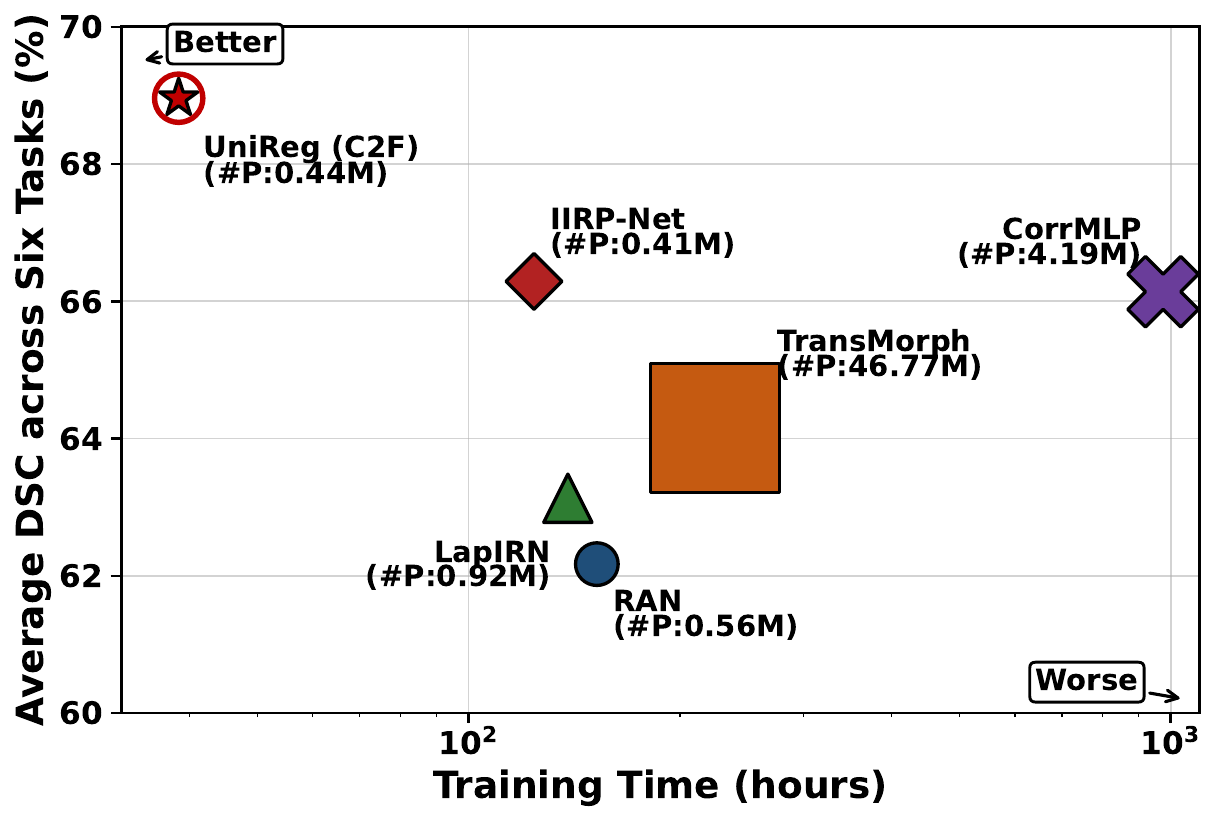}

    \caption{\rev{Efficiency--accuracy comparison of different registration methods. 
The x-axis denotes the total training time, the y-axis denotes the average DSC across six tasks, and the marker size represents the number of model parameters.}}
	\label{fig:Efficiency_accuracy}
\end{figure}

\subsubsection{Baselines}
We compare our method with two popular conventional registration methods, NiftyReg~\cite{SunNK14} and Deeds~\cite{HeinrichJBS12}, six state-of-the-art learning-based approaches, LapIRN~\cite{MokC20}, IIRP-Net~\cite{ma2024iirp}, CorrMLP~\cite{meng2024correlation}, SAME~\cite{LiuYHGLYHXXYJ21}, \rev{RAN~\cite{zheng2024residual}}, \rev{TransMorph~\cite{ChenFHSLD22}}, and one unified model, uniGradICON~\cite{TianGKVEBRN24}. 
We also compare a basic universal training approach for the SAME model: UniSAME, which uses a sequential training process, incorporating HeadNeck, Chest, Abdomen, and Liver datasets. All experiments were conducted on an NVIDIA Tesla $V100$ GPU.

\subsubsection{Implementation Details}
\rev{For UniReg (CNN),} we employ a pre-trained SAM model to extract a $128$-dimensional global embedding and voxel-wise local embeddings. To ensure dimensional compatibility, the global embedding is linearly interpolated to the spatial resolution of the local embeddings. We then apply $L_2$ normalization to the extracted feature embeddings. The resized global feature map, local feature map, and original image are concatenated along the channel dimension and fed into UniReg.

\rev{For UniReg (C2F/MLP), we adopt a two-stage training strategy. 
\revMinor{This is because the Brain MR task empirically benefits from a larger learning rate of $1.0\times10^{-4}$, whereas most other tasks are trained with $1.0\times10^{-5}$. Directly using a single learning rate for all tasks in joint training would therefore be suboptimal. }
In the first stage, the model is trained on the Brain MR task for $80{,}000$ iterations \revMinor{with a learning rate of $1.0\times10^{-4}$. }
In the second stage, the Brain-pretrained model is used to initialize the six-task joint training. During joint training, each optimization iteration samples one task according to the predefined sampling weights, and the model is optimized across all six tasks in a unified manner \revMinor{with a learning rate of $1.0\times10^{-5}$.} The second-stage joint training is conducted for $150{,}000$ iterations in total. }
\revMinor{Specifically, Brain task receives approximately $21{,}000$ additional updates during Stage-2 joint training. For the other tasks, the task-specific update numbers are approximately $21{,}000$ or $43{,}000$, depending on their sampling weights.}
\rev{The code and model have been publicly released at: \url{https://github.com/Alison-brie/UniReg}.}
\rev{For all UniReg instantiations, including UniReg (CNN), UniReg (C2F), and UniReg (MLP), the detailed training configuration for the six-task joint training is summarized in Table~\ref{tab:six_task_training_settings}, including the task ID, sampling weight, learning rate, similarity loss, NCC window size, similarity weight, regularization loss, and regularization weight. Since the Brain MR task is involved in both the pretraining and joint-training stages, its settings are reported separately as Brain (Stage 1) and Brain (Stage 2).}

\begin{table}[t]
\centering
\setlength{\tabcolsep}{4pt}
\renewcommand{\arraystretch}{1.18}
\fontsize{9.2pt}{11.pt}\selectfont
\begin{tabular}{>{\raggedright\arraybackslash}p{1.55cm}
                >{\centering\arraybackslash}p{1.85cm}
                >{\centering\arraybackslash}p{1.85cm}
                l}
\toprule
\multirow{2}{*}{\textbf{Dataset}} & \textbf{Single training} & \textbf{Joint training} & \multirow{2}{*}{\textbf{DSC}} \\
\midrule
\multirow{2}{*}{HeadNeck} 
& $\checkmark$ &  & 53.00 \\
&  & $\checkmark$ & 57.04 \textcolor{green!50!black}{\textbf{(+4.04)}} \\
\midrule
\multirow{2}{*}{Chest}    
& $\checkmark$ &  & 51.62 \\
&  & $\checkmark$ & 56.69 \textcolor{green!50!black}{\textbf{(+5.07)}} \\
\midrule
\multirow{2}{*}{Abdomen}  
& $\checkmark$ &  & 52.01 \\
&  & $\checkmark$ & 55.08 \textcolor{green!50!black}{\textbf{(+3.07)}} \\
\midrule
\multirow{2}{*}{Liver}    
& $\checkmark$ &  & 84.79 \\
&  & $\checkmark$ & 87.08 \textcolor{green!50!black}{\textbf{(+2.29)}} \\
\midrule
\multirow{2}{*}{Cardiac}    
& $\checkmark$ &  & 75.37 \\
&  & $\checkmark$ & 76.10 \textcolor{green!50!black}{\textbf{(+0.73)}} \\
\midrule
\multirow{2}{*}{Brain}    
& $\checkmark$ &  & 81.77 \\
&  & $\checkmark$ & 81.79 \textcolor{green!50!black}{\textbf{(+0.02)}} \\
\bottomrule
\end{tabular}
\caption{\rev{Comparison between single-task training and joint training on different datasets. Both settings use the same backbone and training protocol. DSC denotes the Dice similarity coefficient, and the values in parentheses indicate the improvement over single-task training.}}
\label{tab:single_joint_training}
\end{table}

\begin{table}[t]
\centering
\setlength{\tabcolsep}{4.pt}
\renewcommand{\arraystretch}{1.1}
\begin{tabular}{lccccc}
\toprule
\multirow{2}{*}{\textbf{Method}} 
& \multicolumn{2}{c}{\textbf{Lung GTV}} 
& \multicolumn{2}{c}{\textbf{DIR-QA}} 
& \multirow{2}{*}{\begin{tabular}[c]{@{}c@{}}\textbf{Test time}\\\textbf{(sec)}\end{tabular}} \\
\cmidrule(lr){2-3} \cmidrule(lr){4-5}
& $\mathrm{DSC}_{1}$ 
& $\%(|J_{\phi}|<0)$ 
& $\mathrm{TRE}$ $\downarrow$
& $\%(|J_{\phi}|<0)$ 
& \\
\midrule
Initial        & 67.06 & 0.00  & 10.09  & 0.00  & ---    \\
NiftyReg       & 74.59 & 0.02 &  8.16 & 0.01  & 126.94 \\
Deeds          & 81.53 & 0.49 &  4.10 & 9.38  & 134.82 \\
UniReg (C2F)         & 74.04 & 0.00 & 5.53 & 1.12 & 0.07 \\
\bottomrule
\end{tabular}
\caption{\rev{Quantitative results on unseen tasks. TRE denotes the target registration error between corresponding anatomical landmarks. UniReg achieves comparable performance to strong optimization-based registration methods while substantially reducing test time.}}
\label{tab:unseen_result}
\end{table}

\revMinor{All baselines use the same pre-alignment and/or preprocessing pipeline as our method across all tasks.}
\rev{All learning-based comparison methods are trained from scratch on the six in-distribution tasks. For each task, the number of optimization iterations is set to $80{,}000$, using the same task-specific loss and regularization settings as UniReg. For the Brain MR task, the comparison methods are also trained with a learning rate of $1\times10^{-4}$, consistent with the Brain Stage-1 setting in UniReg. UniGradICON is a foundation model for diffeomorphic registration that has been pre-trained on diverse datasets; therefore, we directly use the official repository without additional retraining. All learning-based methods are trained with a batch size of $1$.}

\begin{figure*}[!t]
	\centering
    \includegraphics[width=0.8\linewidth]{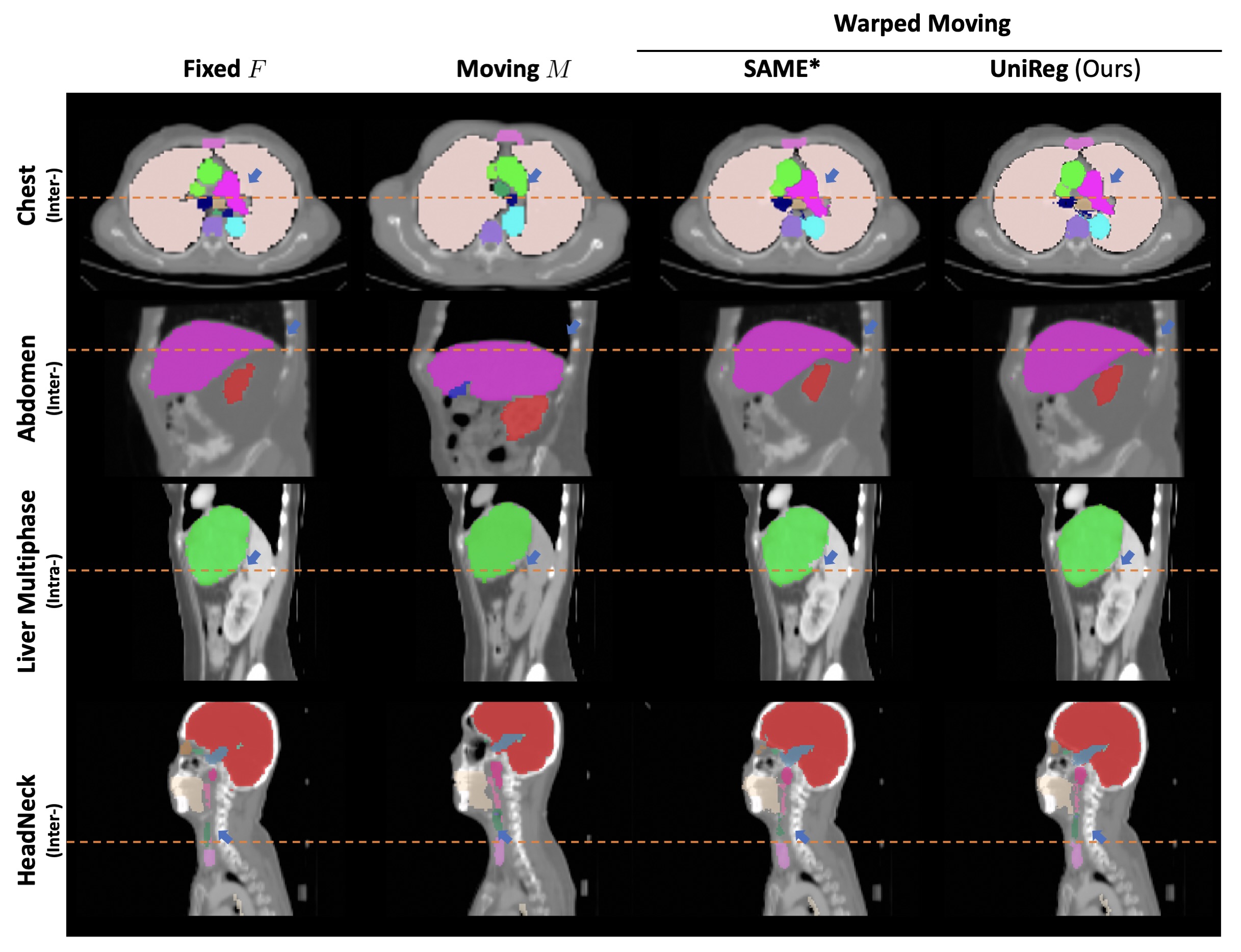}
    \vspace{-0.2em} 
    \caption{ Example slices of the top two registration methods. The warped anatomical segmentations are overlaid and major registration artifacts are highlighted with blue arrows. SAME$*$ denotes the semi-supervised model. \textbf{Chest:} \textit{A.ascendens, A.descendens, A.pulmonary, A.vertebral R, Bronchus R, Lung L, Spine, Sternum, Trachea, V.subclavian R}. \textbf{Abdomen:} \textit{liver and spleen}. \textbf{Liver}: \textit{tumor}. \textbf{HeadNeck:} \textit{Brain, ETbone L, Eye L, Hippocampus L, IAC R, MiddleEar R, OpticNerve R, Pharynx, Thyroid, TympanicCavity L.}
    }
	\label{fig:qualitative}
\end{figure*}

\begin{table*}[t]
\centering
\setlength{\tabcolsep}{5pt}
\renewcommand{\arraystretch}{1.15}
\definecolor{hl}{RGB}{241,184,140}

\resizebox{0.9\textwidth}{!}{
\begin{tabular}{lcccccccccc}
\toprule
\multirow{2}{*}{\textbf{Method}} 
& \multirow{2}{*}{\textbf{With Dice Loss}} 
& \multicolumn{2}{c}{\textbf{HeadNeck CT}} 
& \multicolumn{2}{c}{\textbf{Chest CT}} 
& \multicolumn{2}{c}{\textbf{Abdomen CT}} 
& \multicolumn{2}{c}{\textbf{Liver CT}} 
& \multirow{2}{*}{\textbf{Avg. DSC}} \\
\cmidrule(lr){3-4}
\cmidrule(lr){5-6}
\cmidrule(lr){7-8}
\cmidrule(lr){9-10}
& 
& \textbf{DSC$_{40}\uparrow$} & \textbf{SDlogJ}$\downarrow$
& \textbf{DSC$_{35}$} & \textbf{SDlogJ}
& \textbf{DSC$_{13}$} & \textbf{SDlogJ}
& \textbf{DSC$_2$} & \textbf{$\%(|J_{\phi}|<0)\downarrow$}
& \\
\midrule

UniSAME
& $\checkmark$
& 46.72
& 1.15
& 51.48
& 1.80
& 47.09
& 1.43
& 84.23
& 0.21
& 57.38\\

SAME
& $\checkmark$
& 56.75
& 0.69
& 62.35
& 1.09
& 56.97
& 1.17
& 85.30
& 0.63
& 65.34\\

UniReg (CNN)
& $\checkmark$
& 55.89
& 0.77
& 61.35
& 1.30
& 55.56
& 1.39
& 86.42
& 3.20
& 64.80\\

\bottomrule
\end{tabular}
}
\caption{Comparison of methods on HeadNeck, Chest, Abdomen, and Liver tasks with Dice supervision.}
\label{tab:ablation_same_unireg}
\end{table*}

\begin{figure}[htp]
	\centering
	\setlength{\tabcolsep}{1pt}
	\renewcommand{\arraystretch}{0.75}
	\begin{tabular}{@{}c@{\hspace{0.5pt}}c@{\hspace{0.5pt}}c@{\hspace{0.5pt}}c@{}}
		\includegraphics[width=0.123\textwidth]{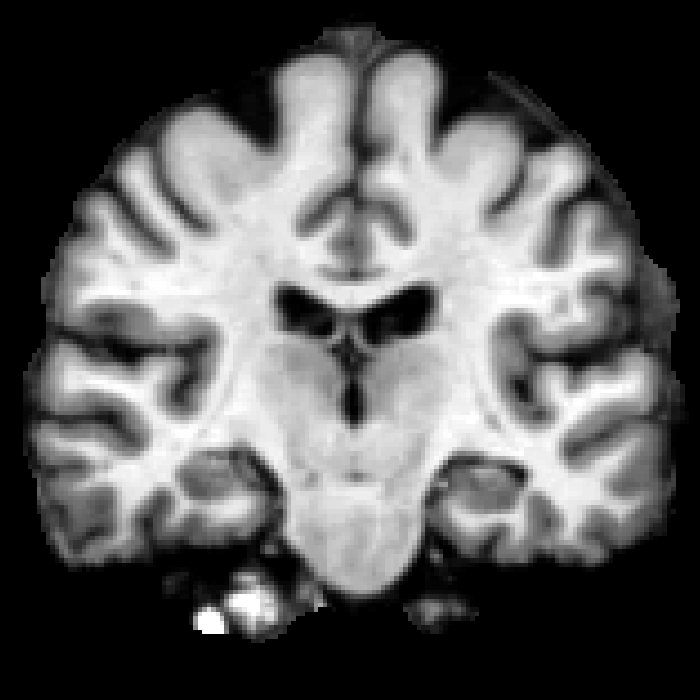}
		&
		\includegraphics[width=0.123\textwidth]{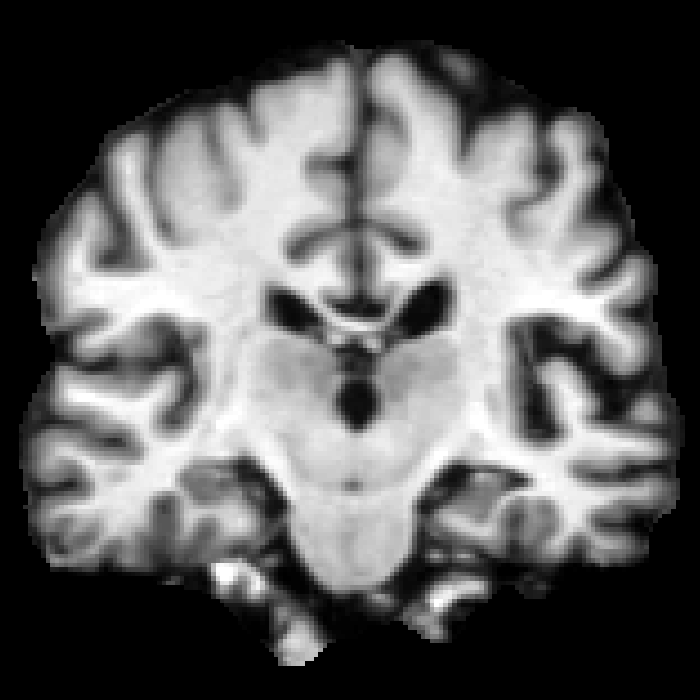}
		&
		\includegraphics[width=0.123\textwidth]{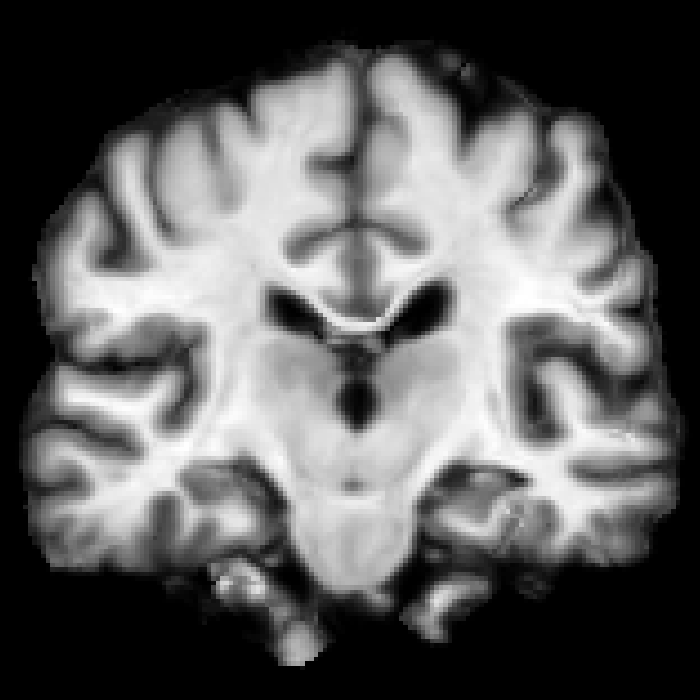}
		&
		\includegraphics[width=0.123\textwidth]{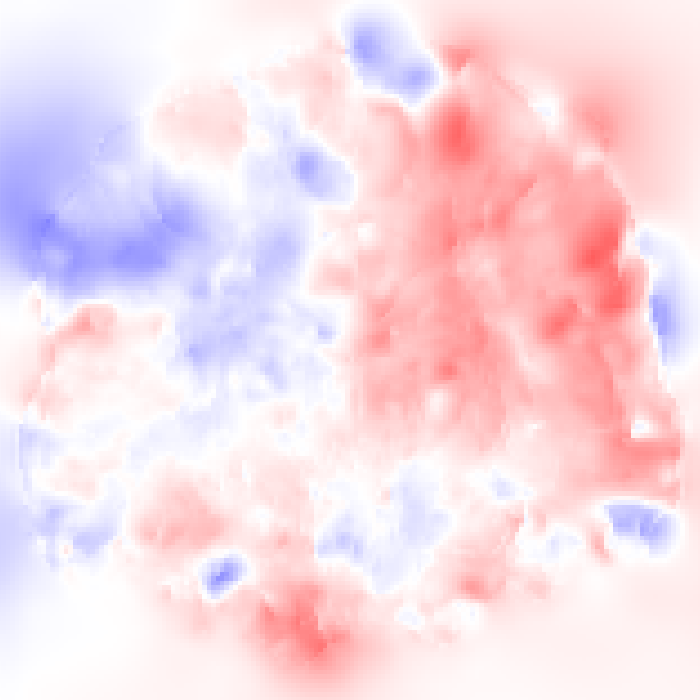}
		\\[-0.25em]
        \includegraphics[width=0.123\textwidth]{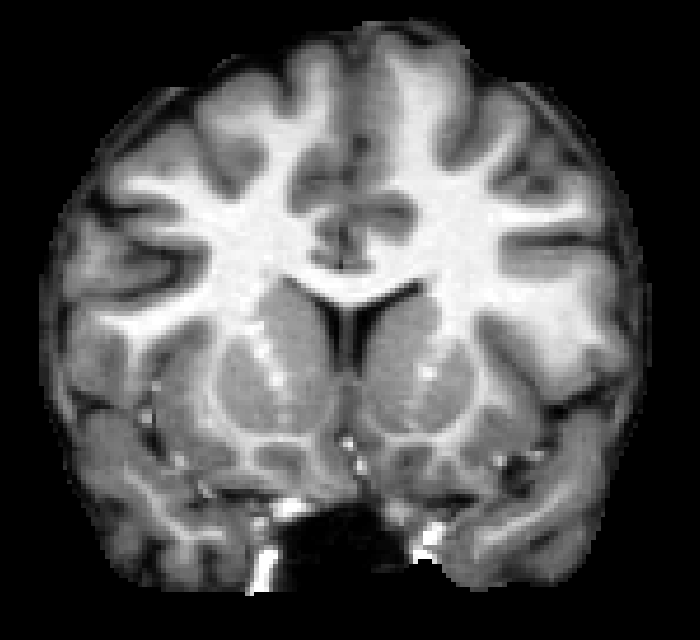}
		&
		\includegraphics[width=0.123\textwidth]{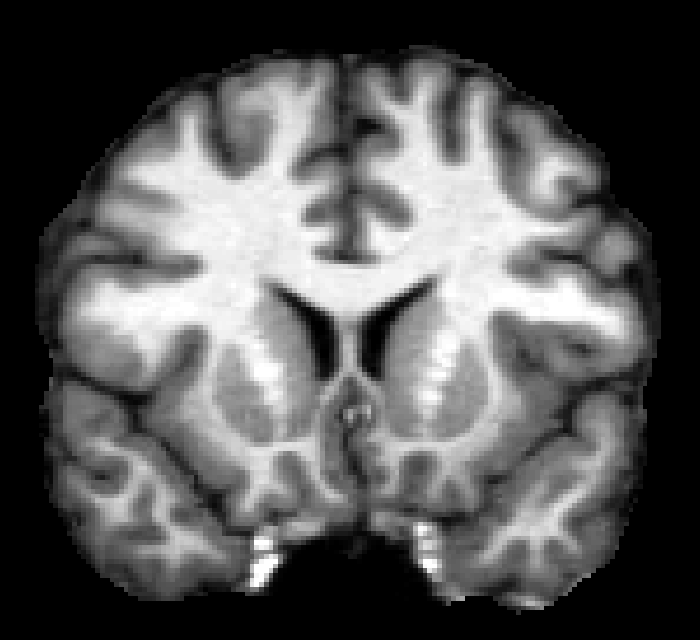}
		&
		\includegraphics[width=0.123\textwidth]{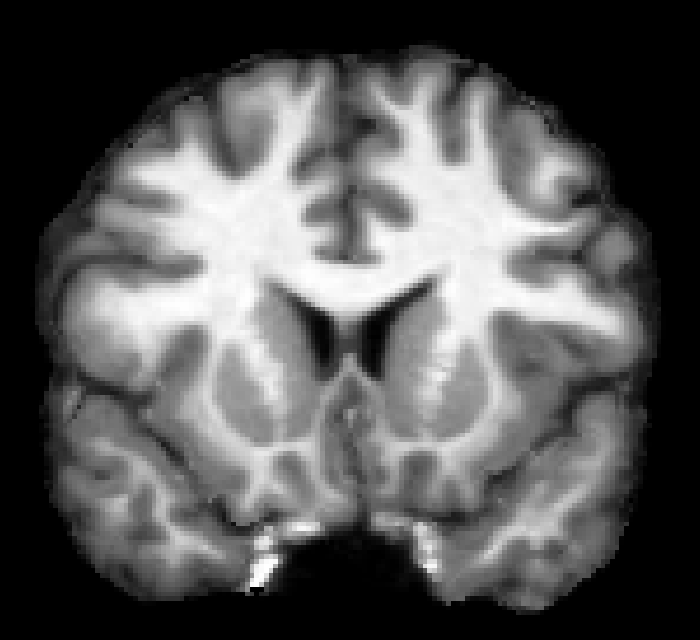}
		&
		\includegraphics[width=0.123\textwidth]{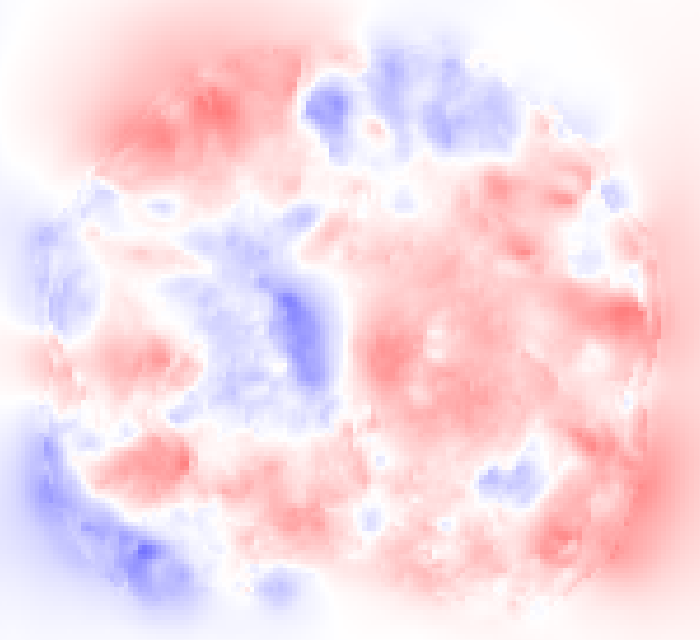}
		\\[-0.25em]
		\scriptsize Moving & \scriptsize Fixed & \scriptsize UniReg & \scriptsize Flow
	\end{tabular}
	\vspace{-0.5em}
	\caption{\rev{Qualitative results on Brain MR registration.}}
	\label{fig:brain_vis}
\end{figure}


\subsection{Comparing to State-of-the-Art Methods}
\label{sec:mainresults}

\subsubsection{Inference on In-distribution Tasks}
Table~\ref{tab:main_result} reports the quantitative comparison between task-specific registration methods and unified models on six in-distribution tasks. To ensure a fair comparison, all learning-based methods are trained without Dice supervision. 

\rev{\textit{Accuracy comparison.}}
UniReg variants achieve consistently competitive and often superior performance across diverse anatomical regions, imaging modalities, and registration settings. Compared with traditional optimization-based methods, learning-based approaches generally provide stronger registration accuracy. Among all the methods, UniReg achieves the best or second-best performance on most tasks. These results demonstrate that the proposed unified registration framework can achieve state-of-the-art or highly competitive accuracy without training separate models for different tasks.
To further examine the effect of segmentation supervision, we report an additional semi-supervised comparison in Table~\ref{tab:ablation_same_unireg}. It shows that UniReg (CNN) also remains competitive when Dice supervision is used. 

\rev{\textit{Flexibility across heterogeneous tasks.}}
Existing task-specific models are usually optimized for a specific anatomical region, imaging modality, or registration type. In contrast, UniReg handles both inter-subject and intra-subject registration within a single unified framework. The evaluated tasks cover CT and MR images, multiple anatomical regions, and different deformation characteristics. As shown in Table~\ref{tab:main_result}, UniReg maintains stable performance across these heterogeneous scenarios, indicating that task-aware conditioning and dynamic deformation generation enable the model to adapt to different registration requirements.

\rev{
\textit{Cross-task benefit.}
To further analyze whether different tasks benefit from each other in the universal model, we added a new comparison between single-task training and joint training, as shown in Table~\ref{tab:single_joint_training}. The results show that joint training consistently improves DSC on all evaluated datasets. These results demonstrate that different tasks can provide complementary anatomical and deformation priors, leading to positive transfer in the universal registration model. }

\rev{
\textit{Efficiency--accuracy trade-off.}
Fig.~\ref{fig:Efficiency_accuracy} further compares learning-based registration methods in terms of average accuracy, total training time, and model size. 
Compared with task-specific learning-based methods, which require training an independent model for each task, UniReg (C2F) achieves the highest average DSC while requiring substantially less total training time and a compact model size. This demonstrates that UniReg provides a favorable trade-off among registration accuracy, training efficiency, and model compactness.}

\textit{Visualization.}  
The qualitative assessments depicted in Figure~\ref{fig:qualitative} provide additional evidence of the robustness and precision of our method in aligning organ and tumor regions. 
\rev{Fig.~\ref{fig:brain_vis} shows representative Brain MR registration results, where UniReg produces well-aligned warped images. }


\subsubsection{Inference on Unseen Tasks}
\rev{
Table~\ref{tab:unseen_result} reports the evaluation results on unseen or out-of-distribution tasks, including Lung GTV registration and Abdominal DIR-QA registration. Most task-specific learning-based baselines cannot be directly transferred in a meaningful way and often fail without retraining. Therefore, we compare UniReg with strong optimization-based registration methods, including NiftyReg and Deeds.
As shown, UniReg (C2F) still achieves reasonable zero-shot registration performance on both unseen tasks. On the Lung GTV task, UniReg improves the DSC from $67.06$ to $74.04$ without introducing folding in the deformation fields. On the DIR-QA benchmark, UniReg reduces the average TRE from $10.09$ mm to $5.53$ mm, outperforming NiftyReg. Although Deeds achieves higher accuracy on the unseen task, it requires substantially longer test time. In contrast, UniReg (C2F) requires only $0.07$ seconds per case, compared with $126.94$ seconds for NiftyReg and $134.82$ seconds for Deeds. These results indicate that UniReg can be directly applied without retraining or instance optimization, while achieving reasonable registration accuracy and substantially faster inference than optimization-based methods.
}

\begin{table}[t]
\centering
\setlength{\tabcolsep}{6pt}
\renewcommand{\arraystretch}{1.18}
\fontsize{9.2pt}{11.pt}\selectfont
\begin{tabular}{>{\raggedright\arraybackslash}p{2.40cm}ccc}
\toprule
\textbf{Organ} & \textbf{$B.\,\lambda$} & \textbf{$B.$ Dice} & \textbf{$M.$ Dice $(\lambda=0.5)$} \\
\midrule
Spleen             & 0.6 & 73.06 & 72.88 \\
Right kidney       & 0.1 & 71.58 & 71.05 \\
Left kidney        & 0.5 & 74.89 & 74.89 \\
Gallbladder        & 0.5 & 42.05 & 42.05 \\
Esophagus          & 0.5 & 42.99 & 42.99 \\
Liver              & 0.6 & 80.13 & 80.00 \\
Stomach            & 0.6 & 34.09 & 33.55 \\
Aorta              & 0.2 & 73.16 & 72.59 \\
Vena cava          & 0.6 & 70.36 & 70.29 \\
Vein               & 0.5 & 31.21 & 31.21 \\
Pancreas           & 0.6 & 27.00 & 26.82 \\
Left adrenal       & 0.6 & 30.82 & 30.58 \\
Right adrenal      & 0.6 & 28.58 & 28.44 \\
\midrule
\textbf{Mean}      & --- & \textbf{52.30} & 52.10 \\
\bottomrule
\end{tabular}
\caption{Comparison between organ-specific and global regularization priors on 13 abdominal organs. 
$B.\,\lambda$ denotes the best organ-specific hyperparameter for each organ, while $M.$ Dice $(\lambda=0.5)$ denotes the result obtained using the best global hyperparameter for the overall abdomen region. Results are obtained using cLapIRN~\cite{mok2021conditional}.}
\label{tab:ab_taskID}
\end{table}

\begin{table}[t]
\centering
\resizebox{0.45\textwidth}{!}{%
\begin{tabular}{lccc}
\toprule
\multirow{2}{*}{\textbf{Model}} & \multirow{2}{*}{\textbf{Iterations}} & \textbf{Abdomen CT} & \textbf{Liver CT} \\
& & \textbf{DSC$_{13}$ } & \textbf{DSC$_{2}$ } \\
\midrule
\multirow{2}{*}{\makecell[l]{Without dynamic\\deformation}} 
& 10k & 42.00 & 80.82 \\
& 40k & 40.86 & 78.36 \\
\midrule
\multirow{3}{*}{\makecell[l]{With dynamic\\deformation}} 
& 10k & 49.61 & 83.87 \\
& 40k & 54.61 & 85.41 \\
& 80k & \textbf{55.56} & \textbf{86.42} \\
\bottomrule
\end{tabular}%
}
\caption{Impact of dynamic deformation generation and training iterations on registration performance.}
\label{tab:ab_con}
\end{table}

\begin{table}[!t]
\centering
\resizebox{0.45\textwidth}{!}{%
\begin{tabular}{cccc}
\toprule
\multirow{2}{*}{\textbf{Task ID}} & \multirow{2}{*}{\textbf{Instance Feature}} & \textbf{Abdomen CT} & \textbf{Liver CT} \\
& & \textbf{DSC$_{13}$} & \textbf{DSC$_{2}$} \\
\midrule
\ding{51} & \ding{55} & \textit{Failed} & \textit{Failed} \\
\ding{55} & \ding{51} & 55.10 & 84.66 \\
\ding{51} & \ding{51} & \textbf{55.56} & \textbf{86.42} \\
\bottomrule
\end{tabular}%
}
\caption{Impact of the correct task ID and instance feature on registration performance.}
\label{tab:ablation_studies}
\end{table}

\begin{table}[!t]
\centering
\setlength{\tabcolsep}{8pt}
\renewcommand{\arraystretch}{1.08}
\resizebox{0.38\textwidth}{!}{%
\begin{tabular}{ccc}
\toprule
\multirow{2}{*}{\textbf{Correct Reg. Type}} 
& \textbf{Abdomen CT} 
& \textbf{Liver CT} \\
& \textbf{DSC$_{13}$} 
& \textbf{DSC$_{2}$ } \\
\midrule
\ding{55} & 53.38 & 86.33 \\
\ding{51} & \textbf{55.08} & \textbf{86.62} \\
\bottomrule
\end{tabular}%
}
\caption{\rev{Impact of the registration-type indicator on registration performance. The anatomical task ID and instance feature are kept unchanged, while the registration-type indicator is either correctly specified or deliberately flipped.}}
\label{tab:regtype_ablation}
\end{table}

\subsection{Ablation Study}
\subsubsection{Task ID Granularity}
We first investigate whether a fine-grained organ-level task definition is necessary for abdominal registration. 
As shown in Table~\ref{tab:ab_taskID}, different abdominal organs may prefer different regularization strengths, indicating the existence of organ-specific deformation characteristics. 
However, using a single region-level hyperparameter for the whole abdomen achieves a comparable mean Dice score to organ-specific tuning, with only a marginal difference of 0.20 Dice points. 
\rev{This result suggests that region-level task encoding provides a reasonable balance between task specificity and model simplicity, avoiding the need for overly fine-grained organ-specific task definitions.}

\subsubsection{Dynamic Deformation Generation}
The dynamic deformation module is conditioned on both input image content and task-specific objectives. As Table~\ref{tab:ab_con} shows, static models (i.e., with fixed kernels) suffer from performance degradation during training, as their fixed optimization targets become misaligned with evolving task demands. In contrast, our dynamic mechanism adaptively reconciles competing objectives, maintaining stable convergence. 

\subsubsection{Conditioning Mechanism Analysis}
We further analyze the contribution of different conditioning signals. Table~\ref{tab:ablation_studies} ablates the conditioning components: misconfigured task IDs lead to a drop in Dice score (e.g., from 86.42 to 84.66 on Liver), underscoring the necessity of accurate conditioning.
Moreover, removing instance-aware features results in failed or unrealistic deformations.
\rev{Finally, we ablate the registration-type indicator while keeping the anatomical task ID and instance feature unchanged. 
As shown in Table~\ref{tab:regtype_ablation}, flipping the registration-type indicator decreases the Dice score on both Abdomen CT and Liver CT, confirming that the correct registration-type condition is also beneficial for generating appropriate deformation fields.}

\begin{figure}[!t]
	\centering
	\setlength{\tabcolsep}{2pt}
	\renewcommand{\arraystretch}{0.88}
	\begin{tabular}{cc}
		\includegraphics[width=0.225\textwidth]{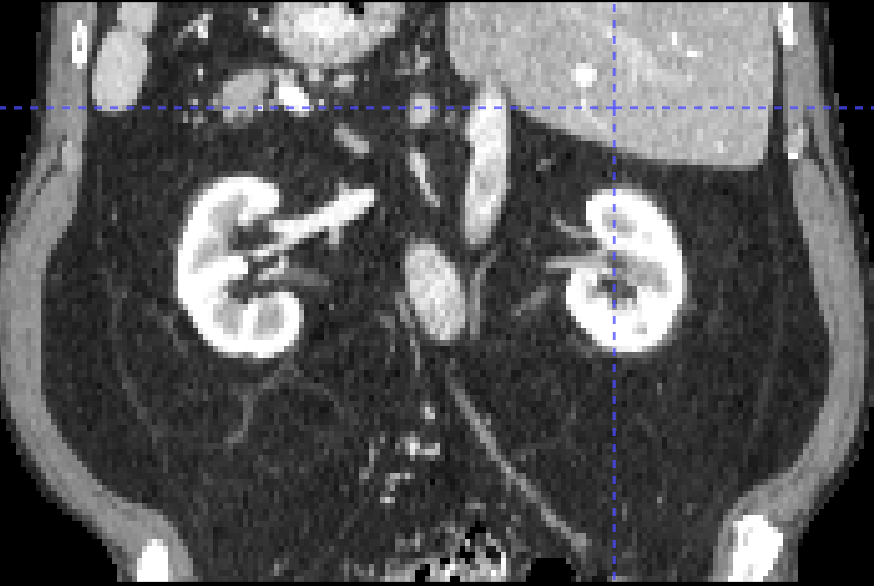}
		&
		\includegraphics[width=0.225\textwidth]{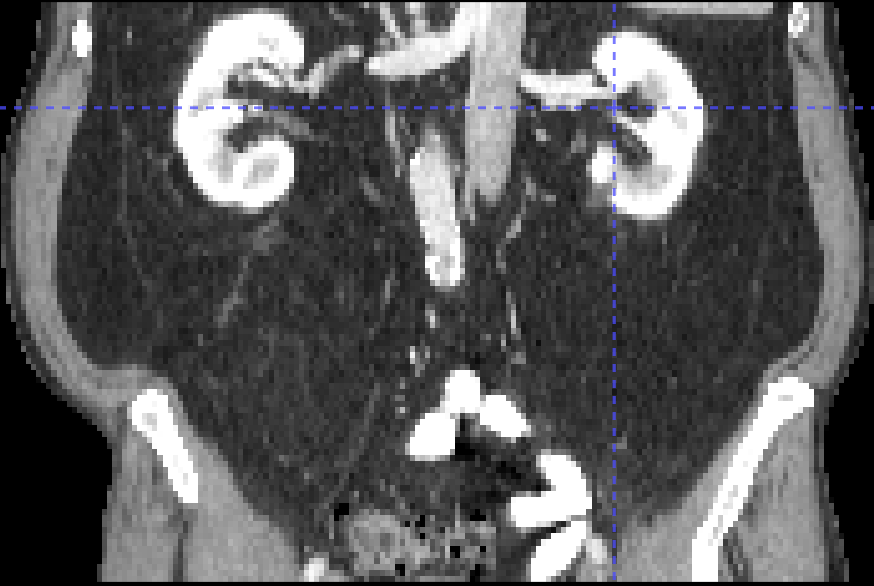}
		\\[-0.35em]
		\small Moving & \small  Fixed 
		\\[0.15em]
		\includegraphics[width=0.225\textwidth]{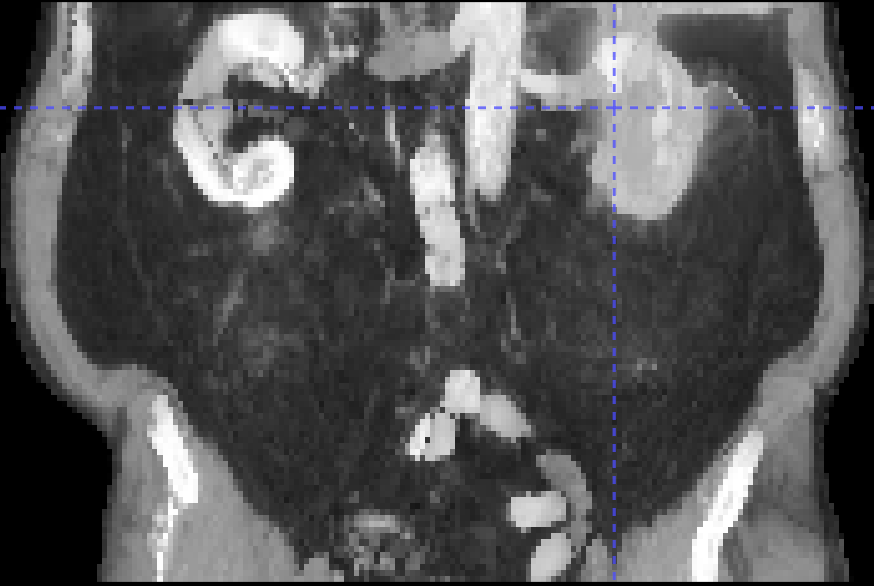}
		&
		\includegraphics[width=0.225\textwidth]{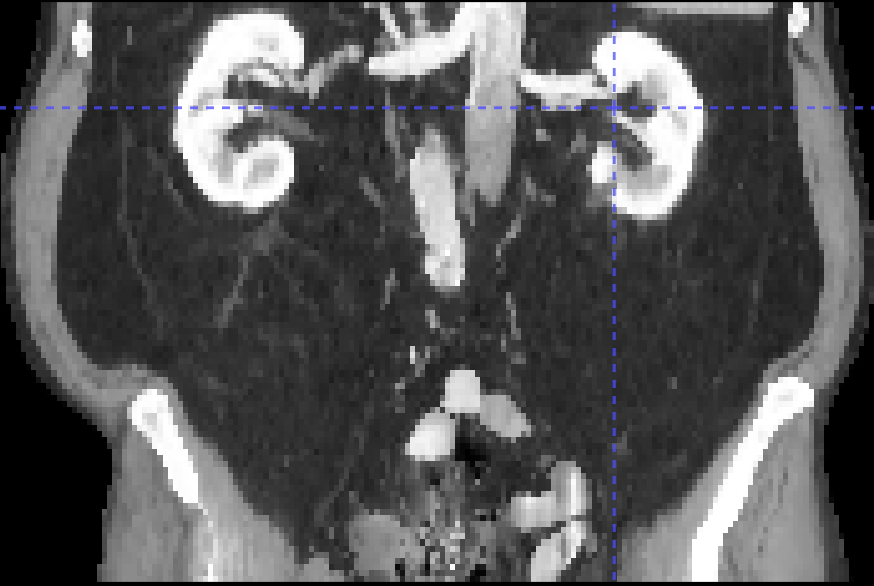}
		\\[-0.35em]
		\small W/O Affine & \small W/ Affine
	\end{tabular}
	\vspace{-0.4em}
	\caption{\rev{Failure case on DIR-QA case15\_Bwd. With severe initial misalignment, UniReg without affine initialization fails, whereas affine pre-alignment enables effective residual deformable refinement.}}
	\label{fig:affine}
	\vspace{-0.6em}
\end{figure}

\subsubsection{\rev{Failure Case Analysis}}
\rev{Fig.~\ref{fig:affine} presents a representative failure case from the DIR-QA dataset, which exhibits severe misalignment with an initial TRE of $51.84$ mm. Directly applying UniReg without affine initialization fails to improve the alignment (TRE = $54.00$ mm), while affine pre-alignment reduces the difficulty to a deformable registration problem and yields substantially better alignment (TRE = $3.69$ mm).}

%% file: sec/5_conclusion.tex
\section{Discussion}

\rev{\textit{Why start from CT imaging?}}
CT is one of the most widely used imaging modalities in radiology and oncology, necessitating precise registration for downstream clinical applications such as diagnosis, treatment planning, and longitudinal monitoring. 
This clinical prevalence motivates our initial focus on scalable CT registration across diverse anatomical regions and registration scenarios.
\rev{\textit{Extension beyond CT.}
Although UniReg was originally motivated by CT registration, its conditional design is not limited to a single modality. 
In this study, we further extend UniReg to brain MR and cardiac MR registration. 
These results demonstrate that the proposed task-aware conditioning and dynamic deformation generation can generalize across heterogeneous modalities.}

\rev{\textit{Scope of universality and limitations of this method.}
We clarify that the term ``universal'' in UniReg does not imply a fully general registration system that can handle all possible imaging modalities, anatomical regions, and deformation patterns without any prior knowledge. 
Rather, it refers to a conditional unified registration framework in which a single model can adapt to multiple registration scenarios through task-aware conditioning and dynamic deformation generation. 
In this study, the evaluated CT datasets alone cover $90$ anatomical structures, including organs and tumors, spanning nearly the entire body from the cranial region to the pelvic area. 
Beyond CT, UniReg is further extended to inter-subject brain MR registration and intra-subject cardiac MR registration, where it achieves strong performance under the same unified framework. 
Moreover, UniReg is evaluated on two unseen tasks, including lung respiratory motion registration and Abdominal DIR-QA, and achieves reasonable zero-shot registration performance without retraining. 
These results demonstrate that UniReg is not limited to a single anatomical region, modality, or registration type, but can scale to heterogeneous CT/MR registration scenarios and unseen registration tasks within a unified model. 
Nevertheless, several limitations remain. 
First, affine or coarse pre-alignment may still be necessary for handling severe global misalignment before deformable registration. 
Second, the current evaluation still covers a limited range of imaging modalities and anatomical scenarios, and its generalization to more challenging cross-modal settings, such as CT--MR or ultrasound--MR registration, remains to be further investigated.}

\section{Conclusion}

We present UniReg, a conditional unified framework for medical image registration. 
By encoding anatomical task identity, registration type, and instance-specific image features into a learnable control vector, UniReg enables a single model to adapt to diverse registration scenarios without requiring multiple task-specific networks. 
The proposed dynamic deformation generation module, together with a knowledge-informed training strategy, enables adaptive alignment across multi-scenario registration tasks. 
\rev{Through comprehensive experiments on multiple CT/MR registration datasets, UniReg achieves superior average registration accuracy compared with current state-of-the-art learning-based methods while exhibiting strong cross-scenario generalization.} 
It also substantially reduces the overall training burden by replacing multiple isolated task-specific models with a compact unified model. 
\rev{Further evaluations with different backbone designs demonstrate the scalability and modality-extensible potential of UniReg.}

%% file: sec/biography.tex
\begin{IEEEbiography}[{\includegraphics[width=1in,height=1.25in,clip,keepaspectratio]{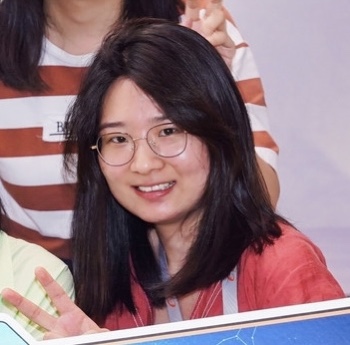}}]{Zi Li} is a PhD candidate at the University of Hong Kong, Hong Kong. Before that, she was an algorithm engineer at Alibaba DAMO Academy. She received the B.E. and M.S.E. degrees from Dalian University of Technology, China. Her research interests focus on medical image registration, segmentation, time-series data analysis, and vision-language pre-training and foundation models.
\end{IEEEbiography}

\begin{IEEEbiography}[{\includegraphics[width=1in,height=1.25in,clip,keepaspectratio]{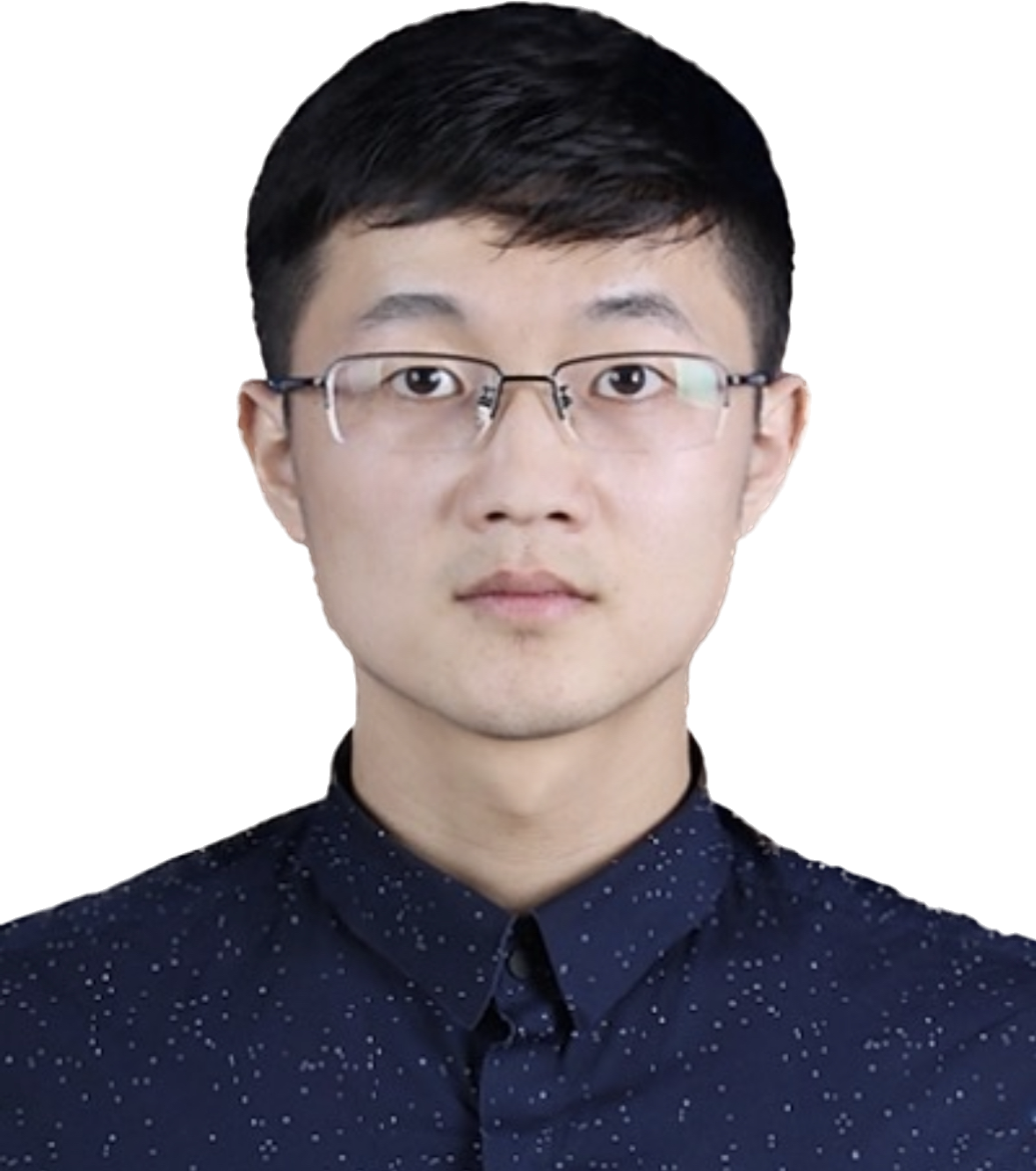}}]{Jianpeng Zhang} is a staff algorithm researcher at Alibaba DAMO Academy, and also a postdoctoral researcher at Zhejiang University.  Before that, he was a research fellow at the Australian Institute for Machine Learning, University of Adelaide, Australia. He received the BE, ME, and PhD degrees from Northwestern Polytechnical University in 2016, 2019, and 2022, respectively. 
His research interests span a broad spectrum in medical AI, including but not limited to multi-modal learning, large language models, and vision-language pre-training. 
\end{IEEEbiography}

\begin{IEEEbiography}[{\includegraphics[width=1in,height=1.25in,clip,keepaspectratio]{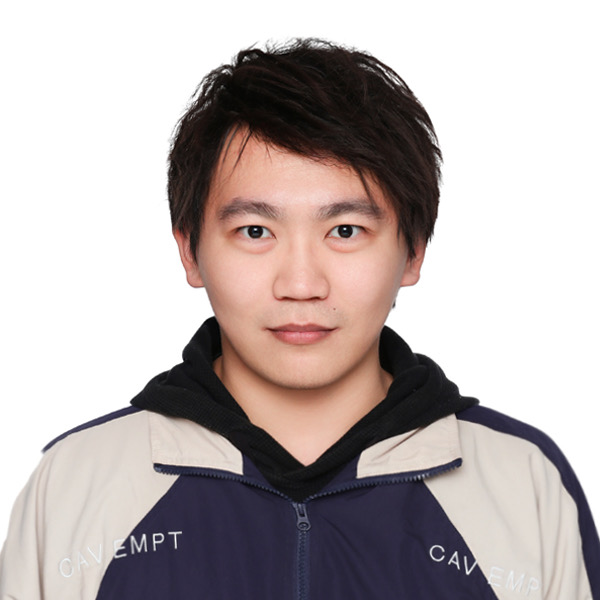}}]{Tai Ma}  received the B.Eng. degree in Computer Science and Technology from East China Normal University (ECNU), Shanghai, China, in 2019, and the Ph.D. degree from ECNU in 2025. Since 2025, he has been with Alibaba DAMO Academy. 
His research interests include image processing and machine learning.
\end{IEEEbiography}

\begin{IEEEbiography}[{\includegraphics[width=1in,height=1.25in,clip,keepaspectratio]{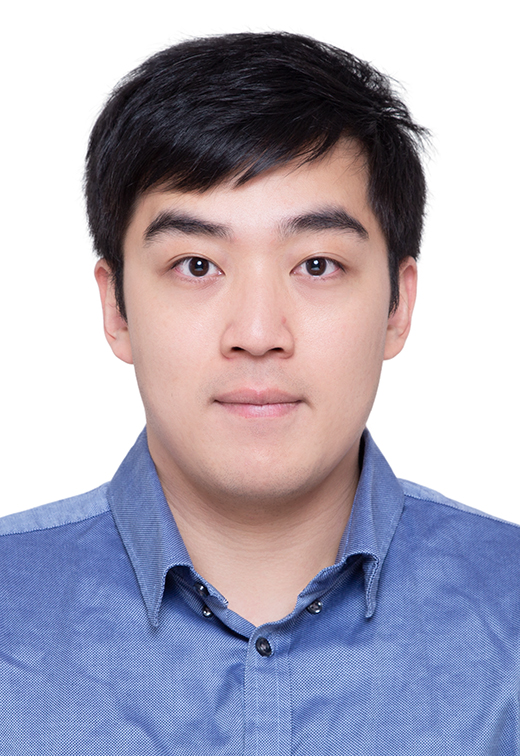}}]{Tony C. W. Mok} is a senior algorithm researcher at DAMO Academy, Alibaba Group. He received the B.Eng. and D.Phil. degrees in computer science from The Hong Kong University of Science and Technology in 2017 and 2022, respectively. 
His research expertise spans several cutting-edge fields, including medical image registration, analysis, computer vision, and deep learning, with a focus on practical applications that improve medical diagnostics and treatment planning.
\end{IEEEbiography}

\begin{IEEEbiography}[{\includegraphics[width=1in,height=1.25in,clip,keepaspectratio]{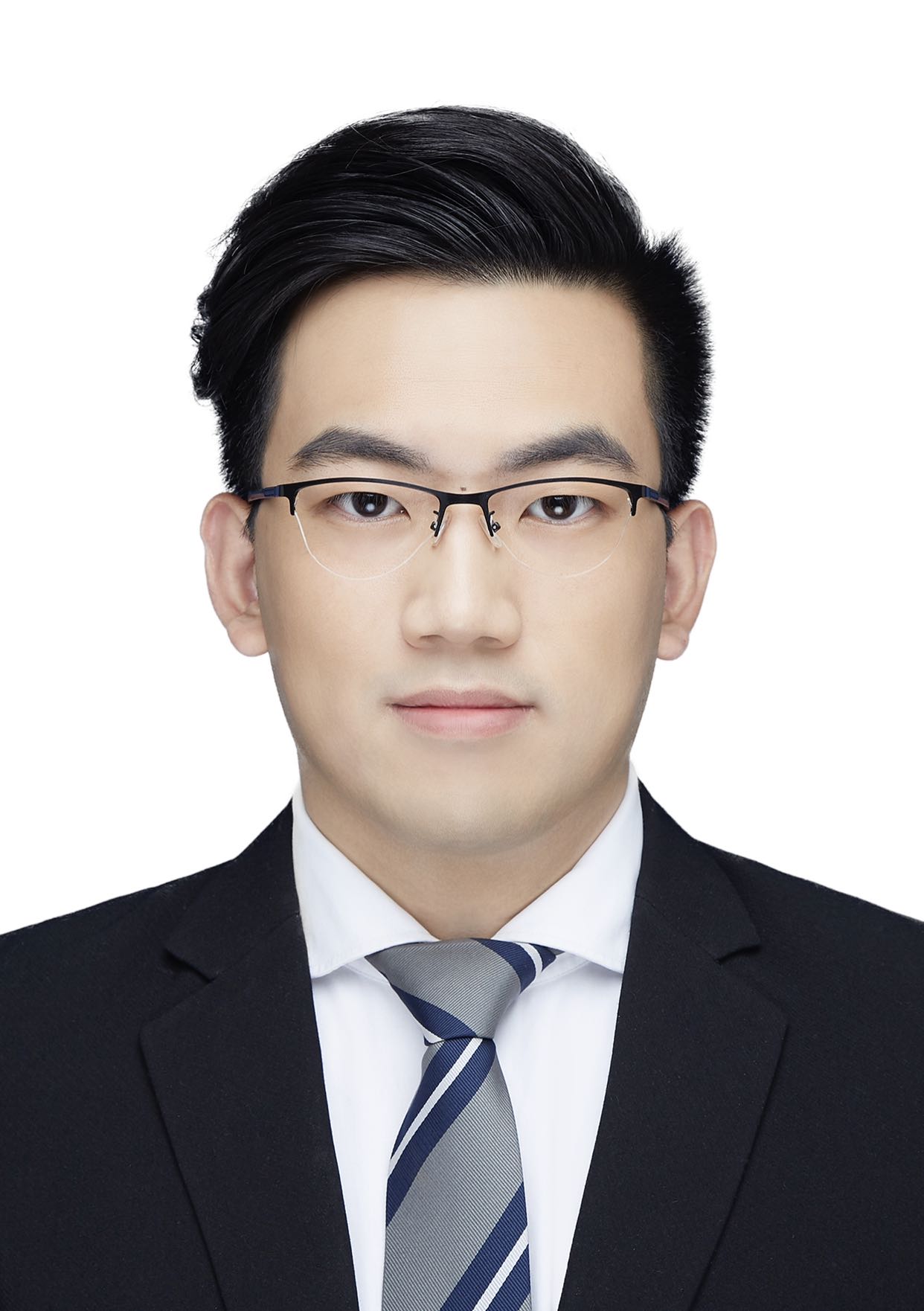}}]{Yan-Jie Zhou} is a senior algorithm researcher at Alibaba DAMO Academy. He is a technical leader in chest pain triple‑rule‑out research at DAMO Academy. He obtained his Ph.D. from the Institute of Automation, Chinese Academy of Sciences. He has published papers and abstracts in Nature Medicine, IEEE Transactions on Medical Imaging, CVPR, MICCAI, ICRA, RSNA, etc. 
His research mainly focuses on robot-assisted intervention and medical image analysis, especially on disease screening and diagnosis in CT images using deep learning.
\end{IEEEbiography}

\begin{IEEEbiography}[{\includegraphics[width=1in,height=1.25in,clip,keepaspectratio]{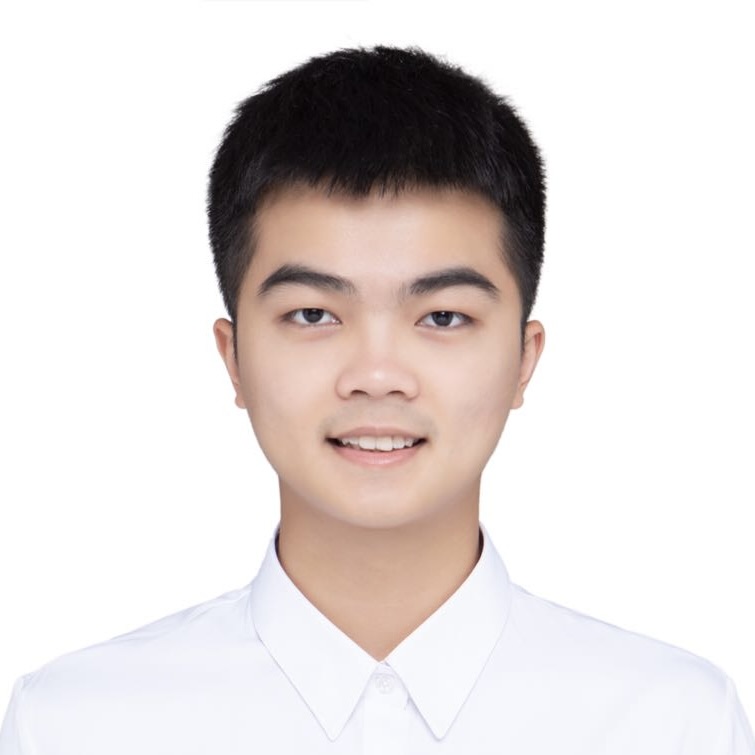}}]{Zeli Chen} is an algorithm engineer at Alibaba DAMO Academy. He received a Master's degree in Biomedical Engineering from Southern Medical University, China, in 2024. His research interests primarily focus on artificial intelligence in medical imaging, including medical image synthesis, medical image segmentation, and radiomics.
\end{IEEEbiography}

\begin{IEEEbiography}[{\includegraphics[width=1in,height=1.25in,clip,keepaspectratio]{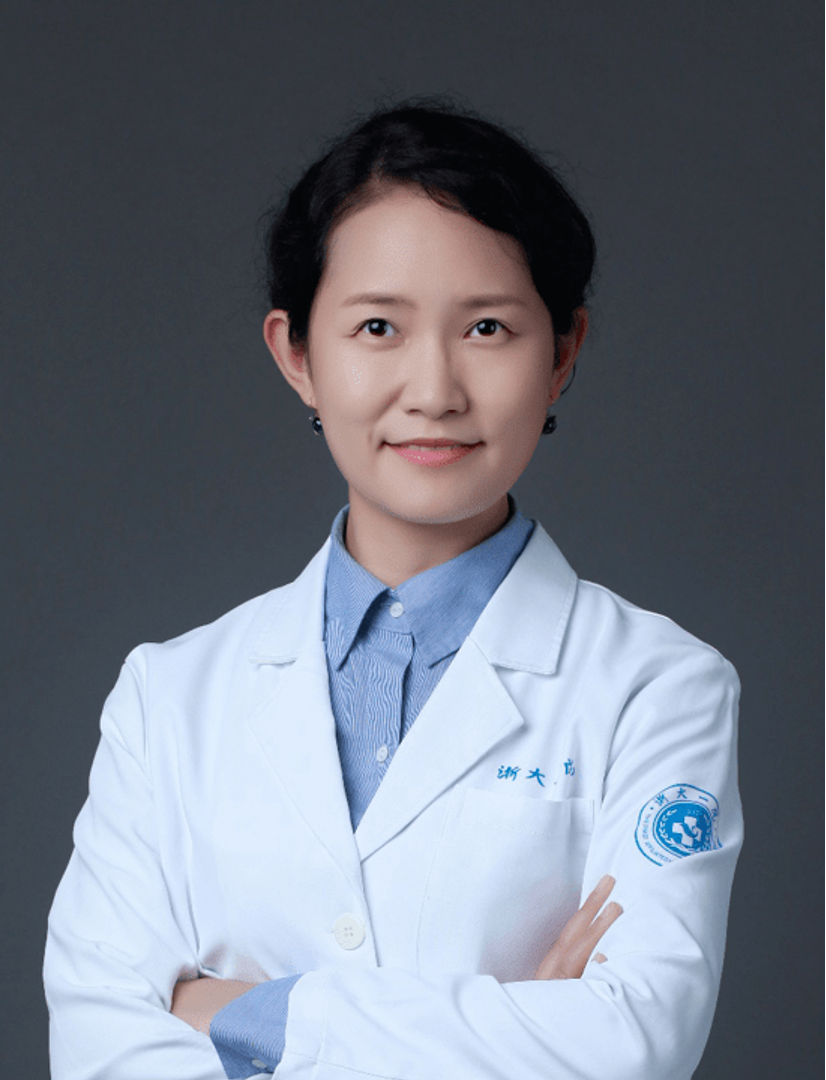}}]{Xianghua Ye} is an Associate Chief Physician and Deputy Director of the Department of Radiation Oncology at the First Affiliated Hospital, School of Medicine, Zhejiang University. She holds an M.D. and a Ph.D. in Radiation Oncology.
She was a visiting scholar at the Stanford University School of Medicine. Dr. Ye has led multiple national and provincial research projects and has authored over 70 peer-reviewed scientific publications in high-impact journals and conferences, including Nature Communications, ICCV, CVPR, MICCAI, and ASTRO. Her primary research interests focus on radiotherapy and chemotherapy for thoracic malignancies and artificial intelligence–assisted diagnosis and treatment in oncology.
\end{IEEEbiography}

\begin{IEEEbiography}[{\includegraphics[width=1in,height=1.25in,clip,keepaspectratio]{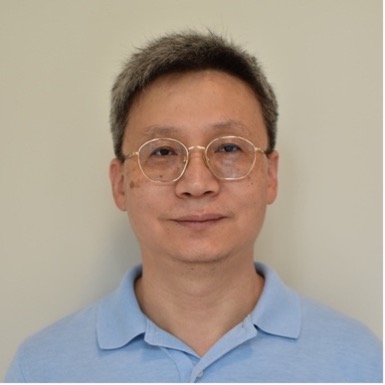}}]{Le Lu} is Head of the Medical \& Health AI Lab at Ant Group and Associate Dean of Ant Group Research Institute since June 2025. Previously, he led the global Medical AI R\&D efforts at Alibaba DAMO Academy. He also worked at PAII Inc., where he led the Bethesda Research Lab, after more than five productive years at the Radiology and Imaging Sciences Department of the Clinical Center, National Institutes of Health. He also worked in NVIDIA’s AI Infrastructure division. From 2006 to 2013, he was a Senior Staff Scientist at Siemens Corporate Research and Siemens Medical Solutions.
Le is an IEEE Fellow for contributions to medical imaging, AI, and oncology imaging, an AIMBE Fellow, a board member of the MICCAI Society, and an IEEE Signal Processing Society Distinguished Industry Speaker. He serves as an Associate Editor for IEEE Transactions on Pattern Analysis and Machine Intelligence, IEEE Transactions on Medical Imaging, IEEE Signal Processing Letters, and Frontiers in Oncology.
Le received his Ph.D. degree in Computer Science from Johns Hopkins University, USA, in 2007.
\end{IEEEbiography}

\begin{IEEEbiography}[{\includegraphics[width=1in,height=1.25in,clip,keepaspectratio]{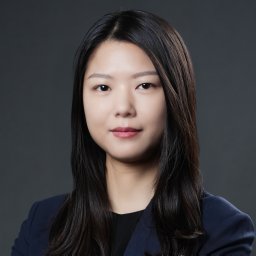}}]{Cheng Chen} is an assistant professor at the Department of Electrical and Computer Engineering, the University of Hong Kong. Before joining HKU, she was a postdoctoral research fellow at Harvard Medical School \& Massachusetts General Hospital. She received her Ph.D. in Computer Science and Engineering from The Chinese University of Hong Kong, her M.S. from The Johns Hopkins University, and her B.S. from Zhejiang University in biomedical engineering.
Her research interests lie in the intersection of artificial intelligence and healthcare, with an emphasis on the application in medical image analysis. The research topics include visual foundation models, cross-modal self-supervised learning, deep model generalization, and robust multi-modal learning.
\end{IEEEbiography}

\begin{IEEEbiography}[{\includegraphics[width=1in,height=1.25in,clip,keepaspectratio]{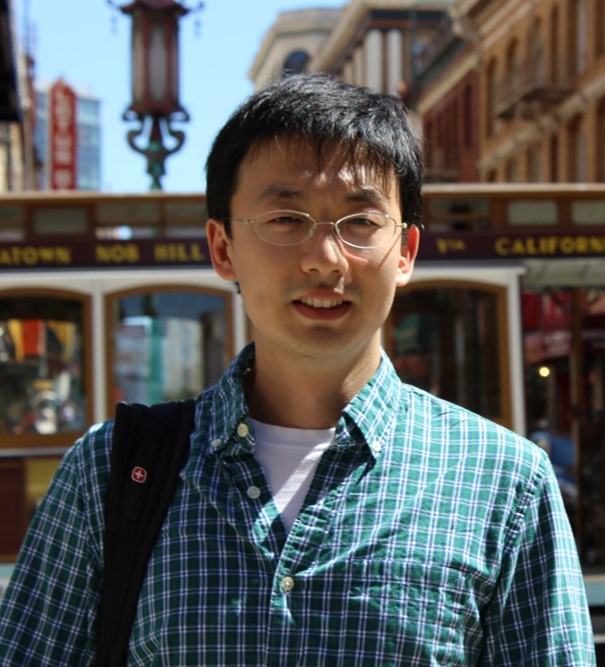}}]{Dakai Jin} is a Staff Algorithm Engineer at Alibaba DAMO Academy, leading a team focused on medical image analysis projects, including radiotherapy target volume segmentation, volumetric image registration, cancer screening, and treatment response prediction. Before joining Alibaba, he was a Staff Scientist at PAII Inc. and a Visiting Research Fellow at the National Institutes of Health. He received his Ph.D. in 2016 from the University of Iowa. 
\end{IEEEbiography}